\documentclass[a4paper,11pt]{article}

\usepackage[utf8]{inputenc} 
\usepackage[T1]{fontenc}    
\usepackage{hyperref}       
\usepackage{url}            
\usepackage{booktabs}       
\usepackage{amsfonts}       
\usepackage{nicefrac}       
\usepackage{microtype}      

\title{One Class Splitting Criteria for Random Forests}

\usepackage{authblk} 
\author[1]{ Nicolas Goix}
\author[2]{ Nicolas Drougard}
\author[1,3]{ Romain Brault}
\author[1]{ Maël Chiapino}
\affil[1]{LTCI, CNRS, Télécom ParisTech, Université Paris-Saclay, Paris, France}
\affil[2]{Institut Supérieur de l'Aéronautique et de l'Espace (ISAE-SUPAERO), Toulouse, France}
\affil[3]{IBISC, Université d'\'Evry val d'Essonne, \'Evry, France}

\usepackage{times}  
\usepackage[utf8]{inputenc} 
\usepackage{amsmath}
\usepackage{amssymb}
\usepackage{dsfont}
\usepackage{caption} 
\usepackage{graphicx}
\usepackage{tikz}
\usepackage{pgfplots}
\usepackage{subcaption}  
\usepackage{booktabs} 

\usepackage{cleveref}
\Crefformat{equation}{(#2#1#3)}
\Crefrangeformat{equation}{(#3#1#4) to~(#5#2#6)}
\Crefmultiformat{equation}{(#2#1#3)}{ and~(#2#1#3)}{, (#2#1#3)}{ and~(#2#1#3)}

\newtheorem{remark}{Remark}

\newcommand{\ie}{\emph{i.e.}{}}

\newcommand{\eg}{\emph{e.g.}{}}
\newcommand\iid{\ensuremath{\mathit{i.i.d.}}\ }
\newcommand{\rv}{\emph{r.v.}{}}
\newcommand{\wrt}{\emph{w.r.t.}{}}

\def\mb{\mathbf}

\def\leb{\text{Leb}}

\newcommand{\vide}[1]{}

\newcommand{\pointSampled}[1]{
    \coordinate (A) at (#1);
    \draw[fill=black] (A) circle (0.03cm);
}

\def\leb{\text{Leb}}
\def\mb{\mathbf}

\date{}

\begin{document}
\maketitle

\begin{abstract}
Random Forests (RFs) are strong machine learning tools for classification and regression. However, they remain supervised algorithms, and no extension of RFs to the one-class setting has been proposed, except for techniques based on second-class sampling.
This work fills this gap by proposing a natural methodology to extend standard splitting criteria to the one-class setting, structurally generalizing RFs to one-class classification.
An extensive benchmark of seven state-of-the-art anomaly detection algorithms is also presented. This empirically demonstrates the relevance of our approach.
\end{abstract}

\section{Introduction}
\label{ocrf:sec:intro}
%
Anomalies, novelties or outliers are usually assumed to lie in low probability regions of the data generating
process.
%
This assumption drives many statistical anomaly detection methods. 
Parametric techniques \cite{Barnett94, Eskin2000} suppose that
the inliers are generated by a distribution belonging to some specific
parametric model a priori known.
Here and hereafter, we denote by inliers the `not abnormal' data, and by outliers/anomalies/novelties the data from the abnormal class.
%
Classical non-parametric approaches are based on density (level set) estimation
\cite{Scholkopf2001, Scott2006, Breunig2000LOF, Quinn2014},
on dimensionality reduction \cite{Shyu2003, Aggarwal2001} or on decision trees
\cite{Liu2008, Shi2012}.
Relevant overviews of current research on anomaly detection can be found in
\cite{Hodge2004survey, Chandola2009survey, Patcha2007survey, Markou2003survey}.

The algorithm proposed in this paper lies in the \emph{novelty detection}
setting, also called 
\emph{one-class classification}. In this framework, we assume that we only observe examples of one
class (referred to as the normal class, or inlier class). The second (hidden) class is called the abnormal class, or outlier class.
The goal is to identify characteristics of the inlier class, such as its
support or some density level sets with levels close to zero.
This setup is for instance used in some (non-parametric) kernel methods such as
One-Class Support Vector Machine (OCSVM) \cite{Scholkopf2001}, which extends
the SVM methodology \cite{Cortes1995,Shawe2004} to handle training using only inliers.
Recently, Least Squares Anomaly Detection (LSAD) \cite{Quinn2014} similarly extends a multi-class probabilistic classifier \cite{Sugiyama2010} to the one-class setting.
%


RFs are strong machine learning tools \cite{Breiman2001}, comparing well with state-of-the-art
methods such as SVM or boosting algorithms \cite{Freund1996}, and used in
a wide range of domains \cite{Svetnik2003, Diaz2006, Genuer2010}.
These estimators fit a number of decision tree
classifiers on different random sub-samples of the dataset.
Each tree is built recursively, according to a splitting criterion based on
some impurity measure of a node.
The prediction is done by an average over each tree prediction.
In classification the averaging is based on a majority vote. 
 Practical and
theoretical insights on RFs are given in \cite{Genuer2008, Biau2008, Louppe2014, Biau2016}.

Yet few attempts have been made to transfer the idea of RFs to one-class
classification \cite{Desir13, Liu2008, Shi2012}.
%
In \cite{Liu2008}, the novel concept of \emph{isolation} is introduced: the Isolation Forest algorithm isolates anomalies, instead of profiling the inlier behavior which is the usual approach. It avoids adapting splitting rules to the one-class setting by using extremely randomized trees, also named extra trees \cite{Geurts2006}: isolation trees are built completely randomly, without any splitting rule. 
Therefore, Isolation Forest is not really based on RFs, the base estimators being extra trees instead of classical decision trees. Isolation Forest performs very well in practice with low memory and time complexities.
In \cite{Desir13, Shi2012}, outliers are generated to artificially form a second
class.
In \cite{Desir13} the authors propose a technique to reduce the number of
outliers needed by shrinking the dimension of the input space. The outliers
are then generated from the reduced space using a distribution complementary to the inlier
distribution. Thus their algorithm artificially generates a second class, to
use classical RFs.
In \cite{Shi2012}, two different outliers generating processes are compared.
In the first one, an artificial second class is created by randomly sampling
from the product of empirical marginal (inlier) distributions. In the second one
outliers are uniformly generated from the hyper-rectangle that contains the
observed data. The first option is claimed to work best in practice, which can
be understood from the curse of dimensionality argument:
in large dimension \cite{Tax2002}, when the outliers distribution is not tightly defined around the target set,
the chance for an outlier to be in the target set becomes very small, so that a huge number of outliers is needed.
%

Looking beyond the RF literature, \cite{Scott2006} proposes a methodology to build dyadic decision trees to estimate minimum-volume sets \cite{Polonik97, Einmahl1992}. This is done by reformulating their structural risk minimization problem to be able to use the algorithm in \cite{Blanchard2004}.
While this methodology can also be used for non-dyadic trees pruning (assuming such a tree has been previously constructed, \eg~using some greedy heuristic), it does not allow to grow such trees.
Also, the theoretical guaranties derived there relies on the dyadic structure assumption.
%
%
In the same spirit, \cite{CLEM14} proposes to use the two-class splitting criterion defined in \cite{Clemencon2009Tree}. This two-class splitting rule aims at producing oriented decision trees with a `left-to-right' structure to address the bipartite ranking task.
Extension to the one-class setting is done by assuming a uniform distribution for the outlier class. Consistency and rate bounds relies also on this left-to-right structure.  
%
%
Thus, these two references \cite{Scott2006, CLEM14} impose constraints on the tree structure (designed to allow a statistical study) which differs then significantly from the general structure of the base estimators in RF. The price to pay is the flexibility of the model, and its ability to capture complex broader patterns or structural characteristics from the data.

In this paper, we make the choice to stick to the RF framework. We do not assume any structure for the binary decision trees. The price to pay is the lack of statistical guaranties -- the consistency of RFs has only been proved  recently \cite{Scornet2015} and in the context of regression additive models. The gain is that we preserve the flexibility and strength of RFs, the algorithm presented here being able to compete well with state-of-the-art anomaly detection algorithms.
Besides, we do not assume any (fixed in advance) outlier distribution as in \cite{CLEM14}, but define it in an adaptive way during the tree building process.

To the best of our knowledge, no algorithm structurally extends (without second class sampling and without alternative base estimators) RFs to one-class classification. 
Here we precisely 
introduce such a methodology. It builds on a natural adaptation of two-class 
splitting criteria to the one-class setting, as well as an adaptation of the two-class majority vote.

\textbf{Basic idea.} To split a node without second class examples (outliers), we proceed as follows.
Each time we look for the best split for a node $t$, we simply replace (in the two-class \emph{impurity decrease} to be maximized
) the second class proportion going to the left child node $t_L$ by the proportion expectation $\leb(t_L)/\leb(t)$ (idem for the right node), $\leb(t)$ being the volume of the rectangular cell corresponding to node $t$.
It ensures that one child node manages to capture the maximum number of observations with a minimal volume, while the other child looks for the opposite. 

This simple idea corresponds to an adaptive modeling of the outlier distribution.
The proportion expectation mentioned above is weighted proportionally to the number of inliers in node $t$. Thus, the resulting outlier distribution is tightly concentrated around the inliers. 
Besides, and this attests the consistency of our approach with the two-class framework, it turns out that the one-class model promoted here corresponds to the asymptotic behavior of an adaptive 
outliers generating methodology.

This paper is structured as follows. \Cref{ocrf:sec:background} provides the reader with necessary background, to address \Cref{ocrf:sec:one-class} which proposes an adaptation of RFs to the one-class setting and describes a generic one-class random forest algorithm. The latter is compared empirically with state-of-the-art anomaly detection methods in \Cref{ocrf:sec:benchmark}. Finally a theoretical justification of the one-class criterion is given in \Cref{sec:ocrf:theory}.
\section{Background on decision trees}\label{ocrf:sec:background}
Let us denote by $\mathcal{X} \subset \mathbb{R}^d$ the $d$-dimensional  hyper-rectangle containing all the observations.
Consider a binary tree on $\mathcal{X}$ whose node values are subsets of $\mathcal{X}$, iteratively produced by splitting $\mathcal{X}$ into two disjoint subsets.
Each internal node $t$ with value $\mathcal{X}_t$ is labeled with a split feature $m_t$ and split value $c_t$ (along that feature), in such a way that it divides $\mathcal{X}_t$ into two disjoint spaces $\mathcal{X}_{t_L} := \{x \in \mathcal{X}_t, x_{m_t} < c_t \}$ and $\mathcal{X}_{t_R} := \{x \in \mathcal{X}_t, x_{m_t} \ge c_t \}$, where $t_L$ (resp. $t_R$) denotes the left (resp. right) children of node $t$, and $x_j$ denotes the $j$th coordinate of vector $x$. Such a binary tree is grown from a sample $ X_1, \ldots,  X_n$ ($\forall i$, $ X_i \in \mathcal{X}$) and its finite depth is determined either by a fixed maximum depth value or by a stopping criterion evaluated on the nodes (\eg~based on an impurity measure).
The external nodes (the \emph{leaves}) form a partition of $\mathcal{X}$.

In a supervised classification setting, these binary trees are called \emph{classification trees} and
prediction is made by assigning to each sample $x \in \mathcal{X}$ the majority class of the leaves containing $x$. This is called the \emph{majority vote}.
Classification trees are usually built using an impurity measure $i(t)$ whose decrease is maximized at each split of a node $t$, yielding an optimal split $(m_t^*, c_t^*)$. The decrease of impurity (also called \emph{goodness of split}) $\Delta i(t, t_L, t_R)$ \wrt~the split $(m_t, c_t)$ and corresponding to the partition $\mathcal{X}_t=\mathcal{X}_{t_L}\sqcup \mathcal{X}_{t_R}$ of the node $t$ is defined as
\begin{align}
\label{ocrf:eq:impurity_measure_decrease}
\Delta i(t, t_L, t_R) = i(t) - p_L i(t_L) - p_R i(t_R),
\end{align}
where $p_L = p_L(t)$ (resp. $p_R = p_R(t)$) is the proportion of samples from $\mathcal{X}_t$ going to $\mathcal{X}_{t_L}$ (resp. to $\mathcal{X}_{t_R}$). The impurity measure $i(t)$ reflects the goodness of node $t$: the smaller $i(t)$, the purer the node $t$ and the better the prediction by majority vote on this node. Usual choices for $i(t)$ are the Gini index \cite{Gini1912} or the Shannon entropy \cite{Shannon2001}.
To produce a randomized tree, these optimization steps are usually partially randomized (conditionally on the data, splits $(m_t^*, c_t^*)$'s become random variables). A classification tree can even be grown totally randomly \cite{Geurts2006}.
In a two-class classification setup, the Gini index is
\begin{align}
\label{ocrf:eq:gini}
  i_G(t) = 2\left(\frac{n_t}{n_t + n_t'}\right) \left( \frac{n_t'}{n_t + n_t'}\right) 
\end{align}
where $n_t$ (resp. $n_t'$) stands for the number of observations with label $0$ (resp. $1$) in node $t$. The Gini index is maximal when $n_t/(n_t + n_t') = n_t'/(n_t + n_t')=0.5$, namely when the conditional probability to have label $0$ given that we are in node $t$ is the same as to have label $0$ unconditionally: the node $t$ does not discriminate at all between the two classes. 

For a node $t$, maximizing the impurity decrease~\Cref{ocrf:eq:impurity_measure_decrease}
is equivalent to minimizing $p_L i(t_L) + p_R i(t_R)$.
Since $p_L = (n_{t_L} + n_{t_L}') / (n_t + n_t')$ and $p_R = (n_{t_R} + n_{t_R}')/(n_t + n_t')$, and the quantity $(n_t + n_t')$ being constant in the optimization problem, this
 is equivalent to minimizing the following proxy of the impurity decrease,
\begin{align}
\label{ocrf:eq:two_class_proxy}
I(t_L, t_R) =   (n_{t_L} + n_{t_L}') i(t_L) + (n_{t_R} + n_{t_R}') i(t_R).
\end{align}
Note that with the Gini index $i_G(t)$ given in \Cref{ocrf:eq:gini}, the corresponding proxy of the impurity decrease is
\begin{align}
\label{ocrf:tc_proxy}
I_G(t_L, t_R)= \frac{n_{t_L} n'_{t_L}}{n_{t_L} +  n'_{t_L}} + \frac{n_{t_R} n'_{t_R}}{n_{t_R} +  n'_{t_R}}.
\end{align}
In the one-class setting, no label is available, hence the impurity measure $i(t)$ 
does not apply to this setup. The standard splitting criterion which consists in minimizing the latter cannot be used anymore.

\section{Adaptation to the one-class setting}
\label{ocrf:sec:one-class}


The two reasons why RFs do not apply to one-class classification are that the standard splitting criterion does not apply to this setup, as well as the majority vote. In this section, we propose a one-class splitting criterion and a one-class version of the majority vote.


\subsection{One-class splitting criterion}
\label{sec:one-class-crit}

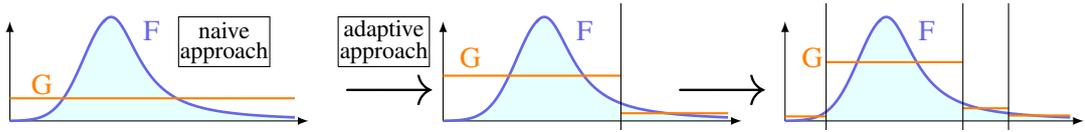
\begin{figure}[!ht]
  \centering
  \begin{tikzpicture}[scale=0.6,declare function={
    c1 = 6/10;
    c2 = 4/10;
    seuil = c1/(c1+c2);
    a1 = 0.1;
    a2 = 0.2;
    x01 = 0.8;
    x02 = 2.9;
    cauchyMass1(\x) = c1*a1/( pi*( pow(a1,2) + pow((\x-x01),2) ));
    cauchyMass2(\x) = c2*a2/( pi*( pow(a2,2) + pow((\x-x02),2) ));
    loinormale(\x) = c2*exp(-pow((\x-x02),2)/0.05);
    loinormale2(\x) = c2*exp(-pow((\x-x02-1.5),2)/0.3);
      indicatorFunction(\x) = exp(-pow(\x-3,2)/6);
      firstVerticalSplitX = 1;
      lastVerticalSplitX = 5.7;
      verticalDashedSplitX = 2.3;
      verticalSplitX = 4;
      lowHorizontalDashedSplitY = 4.5;
      highHorizontalDashedSplitY = 5.5;
      coeffHomothety = 3.8;
      homothetyBone = -1.7;
      homothetyBtwo = -14;
    },
]
  \definecolor{niceblue}{rgb}{0.4,0.4,0.9}
    \definecolor{blue2}{rgb}{0.9,1,1}
\definecolor{ggreen}{rgb}{0.3,0.7,0.4}
\definecolor{orange2}{rgb}{1,0.7,0}


\fill [blue2, domain=-7.9:-1.65, variable=\x]
      (-7.9, 1)
      -- plot[samples=200,smooth] ({\x},{ 1.5*coeffHomothety*loinormale((\x-homothetyBone+15)/coeffHomothety)  +1 } )
      -- (-1.47, 1)
      -- cycle;
\fill [blue2, domain=-5.8:-1.65, variable=\x]
      (-5.85, 1)
      -- plot[samples=200,smooth] ({\x},{ 1.5*0.633*coeffHomothety*cauchyMass2((\x-homothetyBone+15)/coeffHomothety)  +1 } )
      -- (-1.47, 1)
      -- cycle;
  \draw[->,>=latex] (-7.9,1) to (-7.9,3.2);
  \draw[->,>=latex] (-7.9,1) to (-1.35,1);
  \draw [domain=-7.9:-5.8, scale=1, color=niceblue, line width=1pt] plot[samples=200,smooth] (\x,{ 1.5*coeffHomothety*loinormale((\x-homothetyBone+15)/coeffHomothety)  +1} );
  \draw [domain=-5.85:-1.65, scale=1, color=niceblue, line width=1pt] plot[samples=200,smooth] (\x,{ 1.5*0.633*coeffHomothety*cauchyMass2((\x-homothetyBone+15)/coeffHomothety)  +1} );

  \draw[color=orange,thick] (-7.9,1.5) -- (-1.65,1.5);

  \node[color=orange] at (-7.2,1.8){G};
  \node[color=niceblue] at (-4.8,3){F};

  \node at (-3.2,3){\footnotesize naive};
  \node at (-3.2,2.5){\footnotesize approach};
  \draw (-4.2,2.2) -- (-2.2,2.2) -- (-2.2,3.3) -- (-4.2,3.3) -- (-4.2,2.2);

\node at (0.3,3){\footnotesize adaptive};
  \node at (0.3,2.5){\footnotesize approach};
  \draw (-0.7,2.2) -- (1.3,2.2) -- (1.3,3.3) -- (-0.7,3.3) -- (-0.7,2.2);
	\node at (0.4,1.6) {\scalebox{2}{$\longrightarrow$}};


\fill [blue2, domain=1.6:7.95, variable=\x]
      (1.6, 1)
      -- plot[samples=200,smooth] ({\x},{ 1.5*coeffHomothety*loinormale((\x-homothetyBone+5.5)/coeffHomothety)  +1 } )
      -- (8.03, 1)
      -- cycle;
\fill [blue2, domain=3.65:7.85, variable=\x]
      (3.65, 1)
      -- plot[samples=200,smooth] ({\x},{ 1.5*0.633*coeffHomothety*cauchyMass2((\x-homothetyBone+5.5)/coeffHomothety)  +1 } )
      -- (8.03, 1)
      -- cycle;
  \draw[->,>=latex] (1.6,1) to (1.6,3.2);
  \draw[->,>=latex] (1.6,1) to (8.15,1);
  \draw [domain=1.6:3.7, scale=1, color=niceblue, line width=1pt] plot[samples=200,smooth] (\x,{ 1.5*coeffHomothety*loinormale((\x-homothetyBone+5.5)/coeffHomothety)  +1} );
  \draw [domain=3.65:7.85, scale=1, color=niceblue, line width=1pt] plot[samples=200,smooth] (\x,{ 1.5*0.633*coeffHomothety*cauchyMass2((\x-homothetyBone+5.5)/coeffHomothety)  +1} );

\node[color=niceblue] at (4.7,3){F};

\draw[color=orange,thick] (1.6,2) -- (5.5,2);
\draw[color=orange,thick] (5.5,1.17) -- (7.85,1.17);
	 \node[color=orange] at (2.2,2.4){G};

  \draw (5.5,0.8) -- (5.5,3.6);


\fill [blue2, domain=9.1:15.45, variable=\x]
      (9.1, 1)
      -- plot[samples=200,smooth] ({\x},{ 1.5*coeffHomothety*loinormale((\x-homothetyBone-2)/coeffHomothety)  +1 } )
      -- (15.53, 1)
      -- cycle;
\fill [blue2, domain=11.15:15.35, variable=\x]
      (11.15, 1)
      -- plot[samples=200,smooth] ({\x},{ 1.5*0.633*coeffHomothety*cauchyMass2((\x-homothetyBone-2)/coeffHomothety)  +1 } )
      -- (15.53, 1)
      -- cycle;
  \draw[->,>=latex] (9.1,1) to (9.1,3.2);
  \draw[->,>=latex] (9.1,1) to (15.65,1);
  \draw [domain=9.1:11.2, scale=1, color=niceblue, line width=1pt] plot[samples=200,smooth] (\x,{ 1.5*coeffHomothety*loinormale((\x-homothetyBone-2)/coeffHomothety)  +1} );
  \draw [domain=11.15:15.35, scale=1, color=niceblue, line width=1pt] plot[samples=200,smooth] (\x,{ 1.5*0.633*coeffHomothety*cauchyMass2((\x-homothetyBone-2)/coeffHomothety)  +1} );

\node[color=niceblue] at (12.2,3){F};

	\node at (7.7,1.6) {\scalebox{2}{$\longrightarrow$}};

\draw[color=orange,thick] (9.1,1.1) -- (10,1.1);
\draw[color=orange,thick] (10,2.3) -- (13,2.3);
\draw[color=orange,thick] (13,1.28) -- (14,1.28);
\draw[color=orange,thick] (14,1.12) -- (15.35,1.12);
	 \node[color=orange] at (9.7,2.4){G};

  \draw (13,0.8) -- (13,3.6);
\draw (14,0.8) -- (14,3.6);
\draw (10,0.8) -- (10,3.6);

\end{tikzpicture}
  \caption{\small Outliers distribution $G$ in the naive and adaptive approach.  In the naive approach, $G$ does not depends on the tree and is constant on the input space. In the adaptive approach the distribution depends on the inlier distribution $F$ through the tree. The outliers density is constant and equal to the average of $F$ on each node before splitting it.
}
  \label{ocrf:fig:outlier_density}
\end{figure}

As one does not observe the second-class (outliers), $n_t'$ needs to be defined. In the naive approach below, it is defined as $n_t':= n' \leb(\mathcal{X}_t) / \leb(\mathcal{X})$, where $n'$ is the assumed total number of (hidden) outliers. Here and hereafter, $\leb$ denotes the Lebesgue measure on $\mathbb{R}^d$.
In the adaptive approach hereafter, it is defined as $n_t' := \gamma n_t$, with typically $\gamma=1$. Thus, the class ratio $\gamma_t := n_t'/n_t$ is well defined in both approaches and goes to $0$ when $\leb(\mathcal{X}_t) \to 0$ in the naive approach, while it is maintained constant 
to $\gamma$ in the adaptive one.
\paragraph{Naive approach.}
A naive approach to extend the Gini splitting criterion to the one-class setting is to assume a  uniform distribution for the second class (outliers), and to replace their number $n_t'$ in node $t$ by the expectation $n' \leb(\mathcal{X}_t) / \leb(\mathcal{X})$, where $n'$ denotes the total number of outliers (for instance, it can be chosen as a proportion of the number of inliers).
The problem with this approach appears when the dimension is \emph{not small}. As mentioned in the introduction (curse of dimensionality),
when actually generating $n'$ uniform outliers on $\mathcal{X}$, the probability that a node (sufficiently small to yield a good precision) contains at least one of them is very close to zero. That is why data-dependent distributions for the outlier class are often considered \cite{Desir13, Shi2012}.
Taking the expectation $n' \leb(\mathcal{X}_t) / \leb(\mathcal{X})$ to replace the number of points in node $t$ does not solve the curse of dimensionality mentioned in the introduction:
the volume proportion $L_t:=\leb(\mathcal{X}_t) / \leb(\mathcal{X})$ is very close to $0$ for nodes $t$ deep in the tree, especially in large dimension.
In addition, we typically grow trees on sub-samples of the input data, meaning that even the root node of the trees may be very small compared to the hyper-rectangle containing all the input data.
An other problem is that the Gini splitting criterion is skew-sensitive \cite{Flach2003}, and has here to be apply on nodes $t$ with $0 \simeq n_t' \ll n_t$. When trying empirically this approach, we observe that splitting such nodes produces a child containing (almost) all the data (see \Cref{sec:ocrf:theory}).

\textbf{Example}:
To illustrate the fact that the volume proportion $L_t:=\leb(\mathcal{X}_t) / \leb(\mathcal{X})$ becomes very close to zero in large dimension for lots of nodes $t$ (in particular the leaves), suppose for the sake of simplicity that the input space is $\mathcal{X} = [0,1]^d$. Suppose that we are looking for a rough precision of $1/2^3=0.125$ in each dimension, \ie~a unit cube precision of $2^{-3d}$.
To achieve such a precision, the splitting criterion has to be used on nodes/cells $t$ of volume of order $2^{-3d}$, namely with $L_t = 1/2^{3d}$.
Note that if we decide to choose $n'$ to be $2^{3d}$ times larger than the number of inliers in order that $n' L_{t}$ is not negligible \wrt~the number of inliers, the same (reversed) problem of unbalanced classes appears on nodes with small depth.

\paragraph{Adaptive approach.}
Our solution is to remove the uniform assumption on the outliers, and to choose their distribution adaptively in such a way it is tightly concentrated around the inlier distribution. Formally, the idea is to maintain constant the class ratio $\gamma_t := n_t' / n_t$ on each node $t$: before looking for the best split, we update the number of outliers to be equal (up to a scaling constant $\gamma$) to the number of inliers, $n_t' = \gamma n_t$, \ie~$\gamma_t \equiv \gamma$. These (hidden) outliers are uniformly distributed on node $t$. The parameter $\gamma$ is typically set to $\gamma = 1$, see supplementary \Cref{supp:gamma_interpretation} for a discussion on the relevance of this choice (in a nutshell, $\gamma$ has an influence on optimal splits).

\pgfmathsetseed{7}
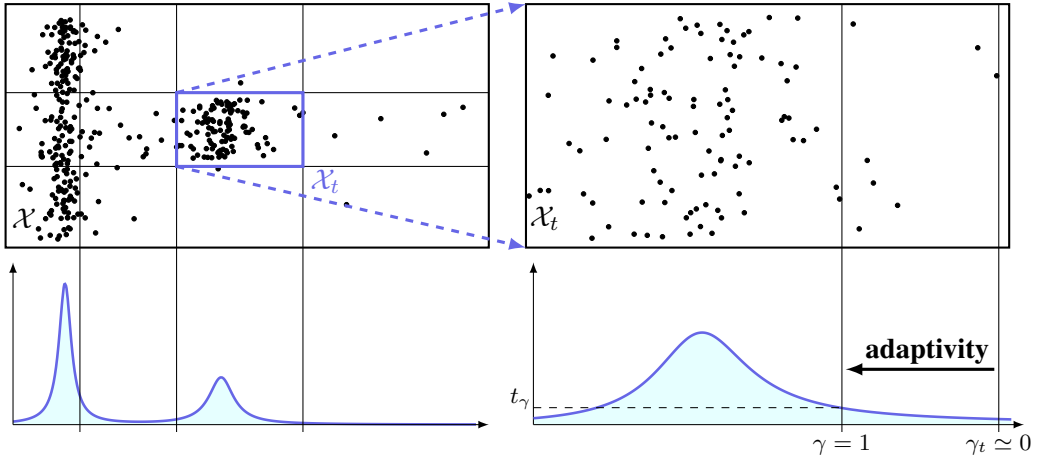
\begin{figure}[ht!]
\center
\resizebox{.9\linewidth}{!} {
\begin{tikzpicture}[declare function={
    c1 = 6/10;
    c2 = 4/10;
    seuil = c1/(c1+c2);
    a1 = 0.1;
    a2 = 0.2;
    x01 = 0.8;
    x02 = 2.9;
    cauchyMass1(\x) = c1*a1/( pi*( pow(a1,2) + pow((\x-x01),2) ));
    cauchyMass2(\x) = c2*a2/( pi*( pow(a2,2) + pow((\x-x02),2) ));
      cauchyRepFuncInv1(\x) = a1*tan( 3.142*(\x-0.5) r) + x01;
      cauchyRepFuncInv2(\x) = a2*tan( 3.142*(\x-0.5) r) + x02;
      indicatorFunction(\x) = exp(-pow(\x-3,2)/6);
      firstVerticalSplitX = 1;
      lastVerticalSplitX = 5.7;
      verticalDashedSplitX = 2.3;
      verticalSplitX = 4;
      lowHorizontalDashedSplitY = 4.5;
      highHorizontalDashedSplitY = 5.5;
      coeffHomothety = 3.8;
      homothetyBone = -1.7;
      homothetyBtwo = -14;
    },
]
  \definecolor{niceblue}{rgb}{0.4,0.4,0.9}
    \definecolor{blue2}{rgb}{0.9,1,1}

    \clip (-0.03,0.5) rectangle (13.8,6.8);

    \draw[thick] (0,3.4) rectangle (6.5,6.7);
    \draw[thick] (7,3.4) rectangle (13.5,6.7);
    \node at (0.25,3.8) {$\mathcal{X}$};
    \node at (7.25,3.8) {$\mathcal{X}_t$};
    \node[color=niceblue] at (verticalSplitX+0.3,lowHorizontalDashedSplitY-0.2) {$\mathcal{X}_t$};

  \foreach \x in {1,2,...,350}{
    \pgfmathsetmacro{\seuil}{c1/(c1+c2)}
    \pgfmathsetmacro{\aleatorio}{rnd}
    \pgfmathsetmacro{\rndCauchy}{\aleatorio>seuil ? 0 : 1 }
    \pgfmathsetmacro{\abscissePoint}{\rndCauchy*cauchyRepFuncInv1(rand) + (1-\rndCauchy)*cauchyRepFuncInv2(rand)}
    \pgfmathsetmacro{\ordinatePoint}{\rndCauchy*(1.5*rand+5) + (1-\rndCauchy)*(rand*0.4+5)}
    \pgfmathsetmacro{\abscissePointFiltered}{ \abscissePoint>6.6 ? -10 : \abscissePoint }
    \pointSampled{\abscissePointFiltered,\ordinatePoint}
    \pgfmathsetmacro{\rightEnough}{\abscissePoint>verticalDashedSplitX ? true : false }
    \pgfmathsetmacro{\leftEnough}{\abscissePoint<verticalSplitX ? true : false }
    \pgfmathsetmacro{\highEnough}{\ordinatePoint>lowHorizontalDashedSplitY ? true : false}
    \pgfmathsetmacro{\lowEnough}{\ordinatePoint<highHorizontalDashedSplitY ? true : false}
    \pgfmathsetmacro{\newAbscisse}{\rightEnough && \leftEnough && \highEnough && \lowEnough ? \abscissePoint*coeffHomothety + homothetyBone : -10  }
    \pointSampled{\newAbscisse,\ordinatePoint*coeffHomothety + homothetyBtwo}
  }

 \fill [blue2, domain=0.1:6.33, variable=\x]
      (0.1, 1)
      -- plot[samples=200,smooth] ({\x},{cauchyMass1(\x) + cauchyMass2(\x) +1} )
      -- (6.33, 1)
      -- cycle;
  \draw [domain=0.1:6.33, scale=1, color=niceblue, line width=1pt, fill=blue2] plot[samples=200,smooth] (\x,{cauchyMass1(\x) + cauchyMass2(\x) +1});
  \draw[->,>=latex] (0.1,1) to (0.1,3.2);
  \draw[->,>=latex] (0.1,1) to (6.5,1);

  \draw (firstVerticalSplitX,0.9) -- (firstVerticalSplitX,6.7); 
  \draw (verticalSplitX,0.9) -- (verticalSplitX,6.7); 

  \draw (verticalDashedSplitX,0.9) -- (verticalDashedSplitX,6.7);
  \draw (0,lowHorizontalDashedSplitY) -- (6.5,lowHorizontalDashedSplitY);
  \draw (0,highHorizontalDashedSplitY) -- (6.5,highHorizontalDashedSplitY);

  \draw[very thick, color=niceblue] (verticalDashedSplitX,lowHorizontalDashedSplitY) rectangle (verticalSplitX,highHorizontalDashedSplitY);
  \draw[very thick, dashed, color=niceblue,->,>=latex] (verticalDashedSplitX,highHorizontalDashedSplitY) -- (7,6.7);
  \draw[very thick, dashed, color=niceblue,->,>=latex] (verticalDashedSplitX,lowHorizontalDashedSplitY) -- (7,3.4);

  \fill [blue2, domain=7.1:13.53, variable=\x]
      (7.1, 1)
      -- plot[samples=200,smooth] ({\x},{ indicatorFunction((\x-7)/5)*coeffHomothety*1.5*cauchyMass2((\x-homothetyBone)/coeffHomothety)  +1 } )
      -- (13.53, 1)
      -- cycle;
  \draw[->,>=latex] (7.1,1) to (7.1,3.2);
  \draw[->,>=latex] (7.1,1) to (13.7,1);
  \draw [domain=7.1:13.53, scale=1, color=niceblue, line width=1pt] plot[samples=200,smooth] (\x,{ indicatorFunction((\x-7)/5)*1.5*coeffHomothety*cauchyMass2((\x-homothetyBone)/coeffHomothety)  +1} );

  \draw (verticalSplitX+7.25,0.9) -- (verticalSplitX+7.25,6.7); 
  \draw (13.36,0.9) -- (13.36,6.7); 
  \node[below] (gammaone) at (verticalSplitX+7.25,1){\footnotesize $\gamma=1$};
  \node[below] (gammat) at (13.36,1){\footnotesize $\gamma_t \simeq 0$};
  \draw[dashed] (7.1,1.23) -- (verticalSplitX+7.25,1.23);
  \node[right] at (6.65,1.35){\footnotesize $t_{\gamma}$};

  \draw[->,>=latex, very thick] (13.3,1.75) to (verticalSplitX+7.3,1.75);
  \node at (verticalSplitX+8.4,2)  {\textbf{adaptivity}};

\end{tikzpicture}
}
\caption{ \small The left part of this figure represents
the dataset under study and the underlying density.
After some splits on this initial node $\mathcal{X}$,
let us consider the node $\mathcal{X}_t$ illustrated in the right part of this figure:
without the proposed adaptive approach, the class ratio
$\gamma_t$ becomes too small
and yields poor splits 
(all the data are in the `inlier side' of the split, which thus does not discriminate at all).
Contrariwise, setting $\gamma$ to one, \ie~using the adaptive approach,
is far preferable.
}
\label{ocrf:fig:split_alpha}

\end{figure}

With this methodology, one cannot derive a one-class version of the Gini index \Cref{ocrf:eq:gini}, but we can define a one-class version of the proxy of the impurity decrease \Cref{ocrf:tc_proxy}, by simply replacing $n_{t_L}'$ (resp. $n_{t_R}'$) by $n_t' \lambda_L$ (resp. $n_t' \lambda_R$), where $\lambda_L := \leb(\mathcal{X}_{t_L}) / \leb(\mathcal{X}_{t})$ and $\lambda_R := \leb(\mathcal{X}_{t_R}) / \leb(\mathcal{X}_{t})$ are the volume proportion of the two child nodes:
\begin{align}
\label{ocrf:oc_proxy_ad2}
I_G^{OC-ad}(t_L, t_R)= \frac{n_{t_L} \gamma n_t \lambda_L}{n_{t_L} + \gamma n_t \lambda_L} + \frac{n_{t_R} \gamma n_t \lambda_R}{n_{t_R} + \gamma n_t \lambda_R}.
\end{align}
Minimization of the one-class Gini improvement proxy \Cref{ocrf:oc_proxy_ad2} is illustrated in \Cref{ocrf:fig:split_alpha}. 
Note that $n_t'\lambda_L$ (resp. $n_t'\lambda_R$) is the expectation of the number of uniform  observations (on $\mathcal{X}_t$) among $n_t'$ (fixed to $n_t' = \gamma n_t$) falling into the left (resp. right) node.

Choosing the split minimizing $I_G^{OC-ad}(t_L, t_R)$ at each step of the tree building process, corresponds to generating $n_t' = \gamma n_t$ outliers each time the best split has to be chosen for node $t$, and then using the classical two-class Gini proxy \Cref{ocrf:tc_proxy}. The only difference is that $n_{t_L}'$ and $n_{t_R}'$ are replaced by their expectations $n_t'\lambda_{t_L}$ and $n_t'\lambda_{t_R}$ in our method.

\textbf{Resulting outlier distribution.}
\Cref{ocrf:fig:outlier_density} shows the corresponding outlier density $G$ (we drop the dependence in the number of splits to keep the notations uncluttered).
Note that $G$ is a piece-wise constant approximation of the inlier distribution $F$. Considering the Neyman-Pearson test $X \sim F$ vs. $X \sim G$ instead of $X \sim F$ vs. $X \sim \text{Unif}$ may seem surprising at first sight. Let us try to give some intuition on why this works in practice. First, there exists (at each step) $\epsilon>0$ such that $G>\epsilon$ on the entire input space, since the density $G$ is constant on each node and equal to the average of $F$ on this node \emph{before splitting it}. If the average of $F$ was estimated to be zero (no inlier in the node), the node would obviously not have been splitted, from where the existence of $\epsilon$.
Thus, at each step, one can also view $G$ as
 a piece-wise approximation of $F_\epsilon := (1 - \epsilon) F + \epsilon \text{Unif}$, which is a mixture of $F$ and the uniform distribution. 
 Yet, one can easily show that optimal tests for the Neyman-Pearson problem $H_0: X \sim F$ vs. $H_1: X \sim F_\epsilon$ are identical to the optimal tests for $H_0: X \sim F$ vs. $H_1: X \sim \text{Unif}$, since the corresponding likelihood ratios are related by a monotone transformation, see \cite{Scott2009} for instance (in fact, this reference shows that these two problems are even equivalent in terms of consistency and rates of convergence of the learning rules). An other intuitive justification is as follows. In the first step, the algorithm tries to discriminate $F$ from $\text{Unif}$. When going deeper in the tree, splits manage to discriminate $F$ from a (more and more accurate) approximation of $F$. Asymptotically, splits become unrelevant since they are trying to discriminate $F$ from itself (a perfect approximation, $\epsilon \to 0$).
%
\begin{remark}({\sc Consistency with the two-class framework})
Consider the following method to generate outliers -- tightly concentrated around the support of the inlier distribution. 
Sample uniformly $n'= \gamma n$ outliers on the rectangular cell containing all the inliers. Split this root node using classical two-class impurity criterion (\eg~minimizing  \Cref{ocrf:tc_proxy}). Apply recursively the three following steps: for each node $t$, remove the potential outliers inside  $\mathcal{X}_t$, re-sample $n_t'=\gamma n_t$ uniform outliers on $\mathcal{X}_t$, and use the latter to find the best split using \Cref{ocrf:tc_proxy}. 
Then, each optimization problem \Cref{ocrf:tc_proxy} we have solved is equivalent (in expectation) to its one-class version \Cref{ocrf:oc_proxy_ad2}.
In other words, by generating outliers adaptively, we can recover (in average) a tree grown using the one-class impurity, from a tree grown using the two-class impurity.
\end{remark}
\begin{remark}({\sc Extension to other impurity criteria})
Our extension to the one-class setting also applies to other impurity criteria. For instance, in the case of the Shannon entropy defined in the two-class setup by
$i_S(t) = \frac{n_t}{n_t + n_t'} \log_2 \frac{n_t + n_t'}{n_t} + \frac{n_t'}{n_t + n_t'} \log_2 \frac{n_t + n_t'}{n_t'},$
the one-class impurity improvement proxy becomes
$I_S^{OC-ad}(t_L, t_R) = n_{t_L} \log_2 \frac{n_{t_L} + \gamma n_t \lambda_L}{n_{t_L}} + n_{t_R} \log_2 \frac{n_{t_R} + \gamma n_t \lambda_R}{n_{t_R}}.$
\end{remark}
\subsection{Prediction: scoring function of the forest}
\label{ocrf:sec:prediction}
 Now that RFs can be grown in the one-class setting using the one-class splitting criterion, the forest has to return a prediction adapted to this framework.
 In other words we also need to extend the concept of majority vote.
\begin{figure}[hb!]
  \centering
  \includegraphics[width=0.6\linewidth]{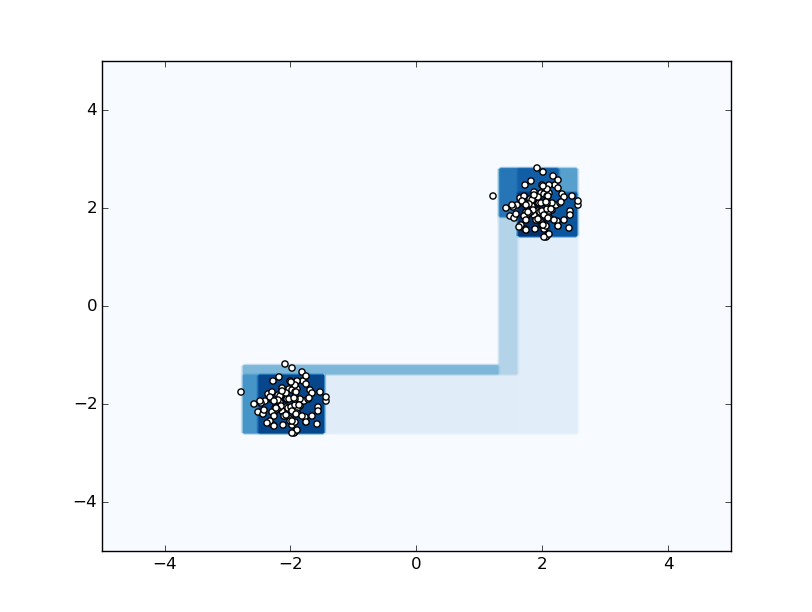}
  \caption{OneClassRF with one tree: level-sets of the scoring function.}
  \label{ocrf:fig:oneclassrf}
\end{figure}
Most usual one-class (or more generally anomaly detection) algorithms actually provide more than just a level-set estimate or a predicted label for any new observation, abnormal vs. normal. Instead, they return a real valued function, termed \textit{scoring function}, defining a pre-order/ranking on the input space. Such a function $s: \mathbb{R}^d \to \mathbb{R}$ allows to rank any observations according to their supposed `degree of abnormality'. Thresholding it provides level-set estimates, as well as a decision rule that splits the input space into inlier/normal and outlier/abnormal regions.
%
%
The scoring function $s(x)$ we use is the one defined in \cite{Liu2008} in view of its established high performance. It is a decreasing function of the average depth of the leaves containing $x$ in the forest. 
An average term is added to each node containing more than one sample, say containing $N$ samples. This term $c(N)$ is the average depth of an extremely randomized tree \cite{Geurts2006} (\ie~built without minimizing any criterion, by randomly choosing one feature and one uniform value over this feature to split on) on $N$ samples. Formally,
\begin{align}
\label{ocrf:eq:scoring3}
\log_2 s(x) = -\left(\sum_{t \text{~leaves}} \mathds{1}_{\{ x \in t \}} d_t + c(n_t)\right) ~/~ c(n),
\end{align}
where $d_t$ is the depth of node $t$, and $c(n) = 2H(n - 1) - 2(n - 1)/n$, $H(i)$ being the harmonic number.
Alternative scoring functions can be defined for this one-class setting (see supplementary \Cref{supp:scoring_functions}).
\subsection{OneClassRF: a Generic One-Class Random Forest algorithm}
Let us summarize the One Class Random Forest algorithm, based on generic RFs \cite{Breiman2001}. It has $6$ parameters:
$max\_samples$, $max\_features\_tree$, $max\_features\_node$, $\gamma$, $max\_depth$, $n\_trees$.
\begin{table*}[ht]
\caption{Original datasets characteristics}
\label{ocrf:table:data}
\centering
\footnotesize
\resizebox{.8\linewidth}{!} {
\begin{tabular}{lccll}
  \toprule
  Datasets        & nb of samples      & nb of features     & ~~~~~~~~~~~~~~~~~~~~~~~~~anomaly class      & ~                  \\ \midrule
  adult       & 48842              & 6                  &    class '$>50K$'                           &      (23.9\%)      \\
  annthyroid  & 7200               & 6                  &    classes $\neq$ 3                         &        (7.42\%)    \\
  arrhythmia  & 452                & 164                &    classes $\neq$ 1 (features 10-14 removed)&  (45.8\%)          \\
  forestcover & 286048             & 10                 &    class 4  (vs. class 2 )                  &           (0.96\%) \\
  http        & 567498             & 3                  &      attack                                 &    (0.39\%)        \\
  ionosphere  & 351                & 32                 &    bad                                      &       (35.9\%)     \\
  pendigits   & 10992              & 16                 &    class 4                                  &        (10.4\%)    \\
  pima        & 768                & 8                  &    pos (class 1)                            &        (34.9\%)    \\
  shuttle     & 85849              & 9                  &      classes $\neq$ 1 (class 4 removed)     &  (7.17\%)          \\
  smtp        & 95156              & 3                  &      attack                                 &    (0.03\%)        \\
  spambase    & 4601               & 57                 &    spam                                     &           (39.4\%) \\
  wilt        & 4839               & 5                  &    class 'w' (diseased trees)               &    (5.39\%)        \\
  \bottomrule
\end{tabular}
}
\end{table*}

Each tree is classically grown on a random subset of both the input samples and the input features \cite{Ho1998, Panov2007}.
This random subset is a sub-sample of size $max\_samples$, with $max\_features\_tree$ variables chosen at random without replacement (replacement is only done after the tree is grown). The tree is built by minimizing \Cref{ocrf:oc_proxy_ad2} for each split, using parameter $\gamma$ (recall that $n_t' := \gamma n_t$), until either the maximal depth $max\_depth$ is achieved or the node contains only one point.
Minimizing \Cref{ocrf:oc_proxy_ad2} is done as introduced in \cite{Amit1997}:
at each node, we search the best split over a random selection of features with fixed size $max\_features\_node$.
The forest is composed of a number $n\_trees$ of trees. The predicted score of a point $x$ is given by $s(x)$, with $s$ defined by \Cref{ocrf:eq:scoring3}.
Remarks on alternative stopping criteria and variable importances are available in supplementary \Cref{supp:stopping_criteria}.

\Cref{ocrf:fig:oneclassrf} represents the level sets of the scoring function produced by OneClassRF, with only one tree ($n\_trees$$=1$) of maximal depth $max\_depth$=4, without sub-sampling, and using the Gini-based one-class splitting criterion with $\gamma=1$.

\section{Benchmarks}
\label{ocrf:sec:benchmark}
In this section, we compare the OneClassRF algorithm described above to seven state-of-art anomaly detection algorithms:
the isolation forest algorithm \cite{Liu2008} (iForest), a one-class RFs algorithm based on sampling a second-class \cite{Desir13} (OCRFsampling), one class SVM \cite{Scholkopf2001} (OCSVM), local outlier factor \cite{Breunig2000LOF} (LOF), Orca \cite{Bay2003}, Least Squares Anomaly Detection \cite{Quinn2014} (LSAD), 
Random Forest Clustering \cite{Shi2012} (RFC).
%
\subsection{Default parameters of OneClassRF}
The default parameters taken for our algorithm are the followings.
$max\_samples$ is fixed to $20\%$ of the training sample size (with a minimum of $100$); $max\_features\_tree$ is fixed to $50\%$ of the total number of features with a minimum of $5$ (\ie~each tree is built on $50\%$ of the total number of features); $max\_features\_node$ is fixed to $5$; $\gamma$ is fixed to $1$; $max\_depth$ is fixed to $\log_2$ (logarithm in base $2$) of the training sample size as in \cite{Liu2008}; $n\_trees$ is fixed to $100$ as in the previous reference. 

The other algorithms in the benchmark are trained with their recommended (default) hyper-parameters as seen
in their respective paper or author's implementation. See supplementary \Cref{supp:hyper_choice} for details.
The characteristics of the twelve reference datasets considered here are summarized
in \Cref{ocrf:table:data}. They are all available on the UCI repository
\cite{Lichman2013} and the preprocessing is done as usually in the litterature (see supplementary \Cref{supp:dataset_description}).
\begin{table*}[ht]
\caption{ Results for the novelty detection setting (novelty detection framework).
}
\label{ocrf:table:results-semisupervised}
\centering
\tabcolsep=0.1cm
\resizebox{1.\linewidth}{!} {
\begin{tabular}{ l  c@{\extracolsep{0.1cm}}c c@{\extracolsep{0.1cm}}c c@{\extracolsep{0.1cm}}c c@{\extracolsep{0.1cm}}c c@{\extracolsep{0.1cm}}c c@{\extracolsep{0.1cm}}c c@{\extracolsep{0.1cm}}c c@{\extracolsep{0.1cm}}c }
\toprule
Datasets & \multicolumn{2}{c }{OneClassRF} & \multicolumn{2}{c }{iForest} & \multicolumn{2}{c }{OCRF\small{sampl.}} & \multicolumn{2}{c }{OCSVM}& \multicolumn{2}{c }{LOF}& \multicolumn{2}{c }{Orca}& \multicolumn{2}{c }{LSAD}& \multicolumn{2}{c }{RFC}  \\
  \cmidrule{1-17}
~     & ROC &  PR & ROC &  PR & ROC & PR  & ROC & PR  & ROC & PR  &ROC  & PR  & ROC &  PR & ROC & PR  \\
adult        &        \textbf{0.665} & \textbf{0.278} & 0.661 & 0.227 & NA & NA & 0.638 & 0.201 & 0.615 & 0.188 & 0.606 & 0.218 &  0.647    & 0.258     & NA & NA \\
annthyroid   &        \textbf{0.936} & 0.468 & 0.913 & 0.456 & 0.918 & \textbf{0.532} & 0.706 & 0.242 & 0.832 & 0.446 & 0.587 & 0.181 &  0.810    & 0.327     & NA & NA \\
arrhythmia   &        0.684 & 0.510 & 0.763 & 0.492 & 0.639 & 0.249 & \textbf{0.922} & \textbf{0.639} & 0.761 & 0.473 & 0.720 & 0.466 &  0.778    & 0.514     & 0.716 & 0.299 \\
forestcover  &        0.968 & 0.457 & 0.863 & 0.046 & NA & NA & NA & NA & \textbf{0.990} & \textbf{0.795} & 0.946 & 0.558 &  0.952    & 0.166     & NA & NA \\
http         &        \textbf{0.999} & \textbf{0.838} & 0.994 & 0.197 & NA & NA & NA & NA & NA & NA & \textbf{0.999} & 0.812 &  0.981    & 0.537     & NA & NA \\
ionosphere   &        0.909 & 0.643 & 0.902 & 0.535 & 0.859 & 0.609 & 0.973 & 0.849 & 0.959 & 0.807 & 0.928 & \textbf{0.910} &  \textbf{0.978}    & 0.893     & 0.950 & 0.754 \\
pendigits    &        0.960 & 0.559 & 0.810 & 0.197 & 0.968 & 0.694 & 0.603 & 0.110 & 0.983 & 0.827 & \textbf{0.993} & \textbf{0.925} &  0.983    & 0.752     & NA & NA \\
pima         &        0.719 & 0.247 & 0.726 & 0.183 & \textbf{0.759} & \textbf{0.266} & 0.716 & 0.237 & 0.700 & 0.152 & 0.588 & 0.175 &  0.713    & 0.216     & 0.506 & 0.090 \\
shuttle      &        \textbf{0.999} & \textbf{0.998} & 0.996 & 0.973 & NA & NA & 0.992 & 0.924 & \textbf{0.999} & 0.995 & 0.890 & 0.782 &  0.996    & 0.956     & NA & NA \\
smtp         &        0.922 & 0.499 & 0.907 & 0.005 & NA & NA & 0.881 & \textbf{0.656} & \textbf{0.924} & 0.149 & 0.782 & 0.142 &  0.877    & 0.381     & NA & NA \\
spambase     &        \textbf{0.850} & 0.373 & 0.824 & 0.372 & 0.797 & \textbf{0.485} & 0.737 & 0.208 & 0.746 & 0.160 & 0.631 & 0.252 &  0.806    & 0.330     & 0.723 & 0.151 \\
wilt         &        0.593 & 0.070 & 0.491 & 0.045 & 0.442 & 0.038 & 0.323 & 0.036 & 0.697 & 0.092 & 0.441 & 0.030 &  0.677    & 0.074     & \textbf{0.896} & \textbf{0.631} \\
\cmidrule{1-17}
average:    & \textbf{0.850} & \textbf{0.495} & 0.821 & 0.311 & 0.769 & 0.410 & 0.749 & 0.410 & 0.837 & 0.462 & 0.759 & 0.454 & \textbf{0.850}  & 0.450 &  0.758  & 0.385 \\
cum. train time: & \multicolumn{2}{c }{\textbf{61s}} & \multicolumn{2}{c }{68s} & \multicolumn{2}{c }{NA} & \multicolumn{2}{c }{NA}& \multicolumn{2}{c }{NA}& \multicolumn{2}{c }{2232s}& \multicolumn{2}{c }{73s}& \multicolumn{2}{c }{NA}  \\

  \bottomrule
\end{tabular}
}
\end{table*}
\subsection{Results}
The experiments are performed in the novelty detection framework, where the training set consists of inliers only. 
%
%
For each algorithm, 10 experiments on random training and testing datasets are
performed, yielding averaged ROC and Precision-Recall curves whose AUCs are
summarized in \Cref{ocrf:table:results-semisupervised} (higher is better).
The training time of each algorithm has been limited (for each experiment among the 10 performed for each dataset) to 30 minutes, where `NA' indicates that the algorithm could not finish training within the allowed time limit.
In average on all the datasets, our proposed algorithm `OneClassRF' achieves both best AUC ROC and AUC PR scores (with LSAD for AUC ROC). It also achieves the lowest cumulative training time.
For further insights on the benchmarks see supplementary \Cref{supp:further_exp}.

%
It appears that OneClassRF has the best performance on five datasets
in terms of ROC AUCs, and is also the best in average.
Computation times (training plus testing) of OneClassRF are also
very competitive. 
%
%
%
%
Experiments in an outlier detection framework (the training set is polluted by
outliers) have also been made (see supplementary \Cref{sup:outlier_detection}). 
The anomaly rate is arbitrarily bounded to $10\%$ max (before splitting data into training and testing sets).
\section{Theoretical analysis} 
\label{sec:ocrf:theory}
This section aims at recovering \Cref{ocrf:oc_proxy_ad2} from a natural modeling of the one-class framework, along with a theoretical study of the problem raised by the naive approach.
\subsection{Underlying model}
\label{ocrf:sec:model}
In order to generalize the two-class framework to the one-class one, we need to consider the population versions associated to empirical quantities \Cref{ocrf:eq:impurity_measure_decrease}, \Cref{ocrf:eq:gini} and \Cref{ocrf:eq:two_class_proxy}, as well as the underlying model assumption. The latter can be described as follows.

\textbf{Existing Two-Class Model (n, $\boldsymbol{\alpha}$).}
We consider a \rv~$X:\Omega \to \mathbb{R}^d$ \wrt~a probability space $(\Omega, \mathcal{F}, \mathbb{P})$.
The law of $X$ depends on another \rv~$y \in \{0,1\}$, verifying $\mathbb{P}(y=1)=1-\mathbb{P}(y=0)=\alpha$. We assume that conditionally on $y=0$, $ X$ follows a law $F$, and conditionally on $y=1$ a law $G$. To summarize:
\begin{align*}
 X ~|~ y=0 ~~\sim~~ F, &~~~~~~~~~~  \mathbb{P}(y=0)=1-\alpha, \\
 X ~|~ y=1 ~~\sim~~ G, &~~~~~~~~~~  \mathbb{P}(y=1)=\alpha.
\end{align*}
Then, considering  $p(t_L | t) = \mathbb{P}( X\in \mathcal{X}_{t_L} |  X\in \mathcal{X}_t)$,  $p(t_R | t) = \mathbb{P}( X\in \mathcal{X}_{t_R} |  X\in \mathcal{X}_t)$, the population version (probabilistic version) of \Cref{ocrf:eq:impurity_measure_decrease} is
\begin{align}
\label{ocrf:eq:impurity_measure_decrease_theo}
\Delta i^{theo}(t, t_L, t_R) ~=~ i^{theo}(t) ~-~  p(t_L | t) i^{theo}(t_L) ~-~  p(t_R | t) i^{theo}(t_R).
\end{align}
It can be used with the Gini index $i_G^{theo}$,
\begin{align}
\label{ocrf:eq:gini_theo}
  i_G^{theo}(t) &~=~ 2 \mathbb{P}(y=0 |  X \in \mathcal{X}_t) \cdot \mathbb{P}(y=1 |  X \in \mathcal{X}_t)
\end{align}
which is the population version of \Cref{ocrf:eq:gini}.

\textbf{One-Class-Model ($n$, $\boldsymbol{\alpha}$).}
We model the one-class framework as follows. Among the $n$ \iid~observations, we only observe those with $y=0$ (the inliers), namely $N$ realizations of $( X ~|~ y=0)$, where $N$ is itself a realization of a \rv~$\mb N$ of law $\mb N \sim \text{Bin}\big(n, (1-\alpha)\big)$. Here and hereafter, $\text{Bin}(n, p)$ denotes the binomial distribution with parameters $(n, p)$. As outliers are not observed, it is natural to assume that $G$ follows a uniform distribution on the hyper-rectangle $\mathcal{X}$ containing all the observations, so that $G$ has a constant density $g(x) \equiv 1 / \leb(\mathcal{X})$ on $\mathcal{X}$. 
Note that this assumption \emph{will be removed} in the adaptive approach described below -- which aims at maintaining a non-negligible proportion of (hidden) outliers in every nodes.

Let us define $L_t=\leb(\mathcal{X}_t)/\leb(\mathcal{X})$. Then, $\mathbb{P}(X \in \mathcal{X}_t,~ y = 1)= \mathbb{P}(y = 1) \mathbb{P}(X \in \mathcal{X}_t|~ y = 1) = \alpha L_t $. Replacing the probability $\mathbb{P}(X \in \mathcal{X}_t, y=0)$ by its empirical version $n_t / n$ in \Cref{ocrf:eq:gini_theo}, we obtain the one-class empirical Gini index
\begin{align}
\label{ocrf:eq:gini_oc}
  i_G^{OC}(t) ~~=~~ \frac{n_t \alpha n L_t}{(n_t + \alpha n L_t)^2}.
\end{align}
This one-class index can be seen as a \emph{semi-empirical} version of \Cref{ocrf:eq:gini_theo}, in the sense that it is obtained by considering empirical quantities for the (observed) inlier behavior and population quantities for the (non-observed) outlier behavior.
Now, maximizing the population version of the impurity decrease $\Delta i_G^{theo}(t, t_L, t_R)$ as defined in \Cref{ocrf:eq:impurity_measure_decrease_theo} is equivalent to minimizing
\begin{align}
\label{ocrf:theo_proxy}
 p(t_L | t)~ i_G^{theo}(t_L) ~+~  p(t_R | t)~ i_G^{theo}(t_R).
\end{align}
Considering semi-empirical versions of $p(t_L | t)$ and $p(t_R | t)$, as for \Cref{ocrf:eq:gini_oc}, gives $p_n(t_L | t) = (n_{t_L} + \alpha n L_{t_L}) / (n_{t} + \alpha n L_{t})$ and $p_n(t_R | t) = (n_{t_R} + \alpha n L_{t_R}) / (n_{t} + \alpha n L_{t})$. Then, the semi-empirical version of \Cref{ocrf:theo_proxy} is
\begin{align}
\label{ocrf:oc_proxy1}
p_n(t_L | t)~ i_G^{OC}(t_L) ~+~  p_n(t_R | t)~ i_G^{OC}(t_R) ~=~ \frac{1}{(n_{t} + \alpha n L_{t})} \left(\frac{n_{t_L}\alpha n L_{t_L}}{n_{t_L} + \alpha n L_{t_L}} + \frac{n_{t_R}\alpha n L_{t_R}}{n_{t_R} + \alpha n L_{t_R}}\right)
\end{align}
where $ 1/(n_{t} + \alpha n L_{t})$ is constant when the split varies.
This means that finding the split minimizing \Cref{ocrf:oc_proxy1} is equivalent to finding the split minimizing
\begin{align}
\label{ocrf:oc_proxy2}
I_G^{OC}(t_L, t_R)= \frac{n_{t_L}\alpha n L_{t_L}}{n_{t_L} + \alpha n L_{t_L}} + \frac{n_{t_R}\alpha n L_{t_R}}{n_{t_R} + \alpha n L_{t_R}}.
\end{align}
Note that \Cref{ocrf:oc_proxy2} can be obtained from the two-class impurity decrease \Cref{ocrf:tc_proxy} as described in the naive approach paragraph in \Cref{ocrf:sec:one-class}. In other words, it is the naive one-class version of \Cref{ocrf:tc_proxy}.
\begin{remark} ({\sc Direct link with the two-class framework})
The two-class proxy of the Gini impurity decrease \Cref{ocrf:tc_proxy} is directly recovered from \Cref{ocrf:oc_proxy2} by replacing $\alpha n L_{t_L}$ (resp. $\alpha n L_{t_R}$) by $n'_{t_L}$ (resp. $n'_{t_R}$), the number of second class instances in $t_L$ (resp. in $t_R$). When generating $\alpha n$ of them uniformly on $\mathcal{X}$, $\alpha n L_{t}$ is the expectation of $n'_{t}$ .
\end{remark}
As detailed in \Cref{sec:one-class-crit}, this approach suffers from the curse of dimensionality.
We can summarize the problem as follows.
Note that when setting $n_t':=\alpha n L_t$, the class ratio $\gamma_t=n_t'/n_t$ is then equal to
\begin{align}
\label{ocrf:def:gamma_t}
\gamma_t = \frac{\alpha n L_t}{n_t}.
\end{align}
This class ratio is close to $0$ for lots of nodes $t$, which makes the Gini criterion unable to discriminate accurately between the (hidden) outliers and the inliers.
%
%
%
Minimizing this criterion produces splits corresponding to $\gamma_t\simeq 0$ in \Cref{ocrf:fig:split_alpha}: one of the two child nodes, say $t_L$ contains almost all the data.
%


\subsection{Adaptive approach}
The solution presented \Cref{ocrf:sec:one-class} is to remove the uniform assumption for the outlier class. From the theoretical point of view, the idea is to choose in an adaptive way (\wrt~the volume of $\mathcal{X}_t$) the number $\alpha n$, which can be interpreted as the number of (hidden) outliers. 
Doing so, we aim at avoiding $\alpha n L_t \ll n_t$ when $L_t$ is too small. Namely, with $\gamma_t$ defined in \Cref{ocrf:def:gamma_t}, we aim at avoiding $\gamma_t \simeq 0$ when $L_t \simeq 0$. The idea is to consider $\alpha(L_t)$ and $n(L_t)$ such that $\alpha(L_t) \to 1$, $n(L_t) \to \infty$ when $L_t \to 0$.
We then define the one-class adaptive proxy of the impurity decrease by
\begin{align}
\label{ocrf:oc_proxy_ad1}
I_G^{OC-ad}(t_L, t_R) ~=~ \frac{n_{t_L}\alpha(L_t) \cdot n(L_t) \cdot L_{t_L}}{n_{t_L} + \alpha(L_t) \cdot n(L_t) \cdot L_{t_L}} ~+~ \frac{n_{t_R}\alpha(L_t) \cdot n(L_t) \cdot L_{t_R}}{n_{t_R} + \alpha(L_t) \cdot n(L_t) \cdot L_{t_R}}.
\end{align}
In other words, instead of considering one general model One-Class-Model($n$, $\alpha$) defined in \Cref{ocrf:sec:model}, we adapt it to each node $t$, considering One-Class-Model($n(L_t)$, $\alpha(L_t)$) \emph{before searching the best split}. We still consider the $N$ inliers as a realization of this model. When growing the tree, using One-Class-Model($n(L_t)$, $\alpha(L_t)$) allows to maintain a non-negligible expected proportion of outliers in the node to be splitted, 
despite $L_t$ becomes close to zero.
Of course, constraints have to be imposed to ensure consistency between these models.
Recalling that the number $N$ of inliers is a realization of $\mb N$ following a Binomial distribution with parameters $(n, 1-\alpha)$, a first natural constraint on $\big(n(L_t), \alpha(L_t)\big)$ is
\begin{align}
\label{ocrf:constraint1}
(1-\alpha)n = \big(1-\alpha(L_t)\big) \cdot n(L_t) \text{~~~~~for all~~} t,
\end{align}
so that the expectation of $\mb N$ remains unchanged. 
\begin{remark}
In our adaptive model One-Class-Model($n(L_t)$, $\alpha(L_t)$) which varies when we grow the tree, let us denote by $\mb N(L_t) \sim \text{Bin}\big(n(L_t), 1-\alpha(L_t)\big)$ the \rv~ruling the number of inliers. The number of inliers $N$ is still viewed as a realization of it. Note that the distribution of $\mb N(L_t)$ converges in distribution to $\mathcal{P}\big((1-\alpha)n\big)$ a Poisson distribution with parameter $(1-\alpha) n$ when $L_t \to 0$, while the distribution $\text{Bin}\big(n(L_t), \alpha(L_t)\big)$ of the \rv~$n(L_t) - \mb N(L_t)$ ruling the number of (hidden) outliers goes to infinity almost surely. In other words, the asymptotic model (when $L_t \to 0$) consists in assuming that the number of inliers $N$ we observed is a realization of $\mb N_\infty \sim \mathcal{P}\big((1-\alpha)n\big)$, and that an infinite number of outliers have been hidden.
\end{remark}
A second natural constraint on $\big(\alpha(L_t), n(L_t)\big)$
is related to the class ratio $\gamma_t$. 
As explained in \Cref{sec:one-class-crit}, we do not want $\gamma_t$ to go to zero when $L_t$ does.
Let us say we want $\gamma_t$ to be constant for all node $t$, equal to $\gamma>0$. From the constraint $\gamma_t = \gamma$ and \Cref{ocrf:def:gamma_t}, we get 
\begin{align}
\label{ocrf:constraint2}
\alpha(L_t) \cdot n(L_t) \cdot L_t 
  = \gamma n_t:=n_t'.
\end{align}
The constant $\gamma$ is a parameter ruling the expected proportion of outliers in each node. Typically, $\gamma=1$ so that there is as much expected uniform (hidden) outliers than inliers at each time we want to find the best split minimizing~\Cref{ocrf:oc_proxy_ad1}.
%
Equations \Cref{ocrf:constraint1} and \Cref{ocrf:constraint2} allow to explicitly determine $\alpha(L_t)$ and $n(L_t)$: $\alpha(L_t) = n_t'/\big((1-\alpha)nL_t + n_t'\big)$ and $n(L_t) = \big((1-\alpha)nL_t + n_t'\big)/L_t$.
Regarding \Cref{ocrf:oc_proxy_ad1}, $\alpha(L_t) \cdot n(L_t) \cdot L_{t_L} = \frac{n_t'}{L_t} L_{t_L} = n_t'\frac{\leb(\mathcal{X}_{t_L})}{\leb(\mathcal{X}_{t})}$ by \Cref{ocrf:constraint2} and $\alpha(L_t) \cdot n(L_t) \cdot L_{t_R}  = n_t'\frac{\leb(\mathcal{X}_{t_R})}{\leb(\mathcal{X}_{t})},$ so that
we recover \Cref{ocrf:oc_proxy_ad2}.
\section{Conclusion}
Through a natural adaptation of (two-class) splitting criteria, this paper introduces a methodology to structurally extend RFs to the one-class setting.
Our one-class splitting criteria correspond to the asymptotic behavior of an adaptive outliers generating methodology, so that consistency with two-class RFs seems respected.
While no statistical guaranties have been derived in this paper, a strong empirical performance attests the relevance of this methodology.

{\small
\bibliographystyle{apalike}
\bibliography{mvextrem}
}


\clearpage
\begin{center}
\textbf{\large Supplementary Materials}
\end{center}
\setcounter{section}{0}
\setcounter{subsection}{0}
\setcounter{equation}{0}
\setcounter{figure}{0}
\setcounter{table}{0}
\setcounter{page}{1}
\makeatletter
\renewcommand\thesection{\Alph{section}}
\renewcommand\thesubsection{\thesection.\arabic{subsection}}
\renewcommand{\theequation}{S\arabic{equation}}
\renewcommand{\thefigure}{S\arabic{figure}}
\renewcommand{\thetable}{S\arabic{table}}

\section{Further insights on the algorithm}\label{supp:further_exp}

\subsection{Interpretation of parameter gamma.}\label{supp:gamma_interpretation}
\begin{figure}
\center
\begin{tikzpicture}[scale=1,declare function={
    c1 = 6/10;
    c2 = 4/10;
    seuil = c1/(c1+c2);
    a1 = 0.1;
    a2 = 0.2;
    x01 = 0.8;
    x02 = 2.9;
    cauchyMass1(\x) = c1*a1/( pi*( pow(a1,2) + pow((\x-x01),2) ));
    cauchyMass2(\x) = c2*a2/( pi*( pow(a2,2) + pow((\x-x02),2) ));
      cauchyRepFuncInv1(\x) = a1*tan( 3.142*(\x-0.5) r) + x01;
      cauchyRepFuncInv2(\x) = a2*tan( 3.142*(\x-0.5) r) + x02;
      indicatorFunction(\x) = exp(-pow(\x-3,2)/6);
      firstVerticalSplitX = 1;
      lastVerticalSplitX = 5.7;
      verticalDashedSplitX = 2.3;
      verticalSplitX = 4;
      lowHorizontalDashedSplitY = 4.5;
      highHorizontalDashedSplitY = 5.5;
      coeffHomothety = 3.8;
      homothetyBone = -1.7;
      homothetyBtwo = -14;
    },
]
  \definecolor{niceblue}{rgb}{0.4,0.4,0.9}
    \definecolor{blue2}{rgb}{0.9,1,1}

    \clip (-0.03,0.5) rectangle (7,6.8);

    \draw[thick] (0,3.4) rectangle (6.5,6.7);
    \node at (0.25,3.8) {$\mathcal{X}$};

  \foreach \x in {1,2,...,350}{
    \pgfmathsetmacro{\seuil}{c1/(c1+c2)}
    \pgfmathsetmacro{\aleatorio}{rnd}
    \pgfmathsetmacro{\rndCauchy}{\aleatorio>seuil ? 0 : 1 }
    \pgfmathsetmacro{\abscissePoint}{\rndCauchy*cauchyRepFuncInv1(rand) + (1-\rndCauchy)*cauchyRepFuncInv2(rand)}
    \pgfmathsetmacro{\ordinatePoint}{\rndCauchy*(1.5*rand+5) + (1-\rndCauchy)*(rand*0.4+5)}
    \pgfmathsetmacro{\abscissePointFiltered}{ \abscissePoint>6.6 ? -10 : \abscissePoint }
    \pointSampled{\abscissePointFiltered,\ordinatePoint}
  }

 \fill [blue2, domain=0.1:6.33, variable=\x]
      (0.1, 1)
      -- plot[samples=200,smooth] ({\x},{cauchyMass1(\x) + cauchyMass2(\x) +1} )
      -- (6.33, 1)
      -- cycle;
  \draw [domain=0.1:6.33, scale=1, color=niceblue, line width=1pt, fill=blue2] plot[samples=200,smooth] (\x,{cauchyMass1(\x) + cauchyMass2(\x) +1});
  \draw[->,>=latex] (0.1,1) to (0.1,3.2);
  \draw[->,>=latex] (0.1,1) to (6.5,1);

  \draw (firstVerticalSplitX,0.9) -- (firstVerticalSplitX,6.7); 
  \draw (verticalSplitX,0.9) -- (verticalSplitX,6.7); 
  \draw (lastVerticalSplitX,0.9) -- (lastVerticalSplitX,6.7); 
  \node[below] at (firstVerticalSplitX,1){\footnotesize $\gamma=10$};
  \node[below] at (verticalSplitX,1){\footnotesize $\gamma=1$};
  \node[below] at (lastVerticalSplitX+0.5,1){\footnotesize $\gamma=0.1$};

\end{tikzpicture}

\caption{Illustration of the standard splitting
criterion on two modes when the proportion $\gamma$ varies.}
\label{ocrf:fig:split_alpha_2}
\end{figure}
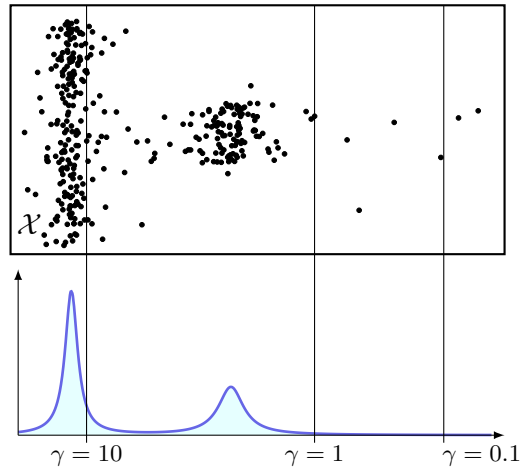

In order for the splitting criterion \Cref{ocrf:oc_proxy_ad2} to perform well, $n_t'$ is expected to be of the same order of magnitude as the number of inliers $n_t$. If $\gamma = n_t'/n_t \ll 1$, the split puts every inliers on the same side, even the ones which are far in the tail of the distribution, thus widely over-estimating the support of inliers. If $\gamma \gg 1$, the opposite effect happens, yielding an estimate of a $t$-level set with $t$ not close enough to $0$.
\Cref{ocrf:fig:split_alpha}
 illustrates the splitting criterion when $\gamma$ varies. It clearly shows that there is a link between parameter $\gamma$ and the level $t_\gamma$ of the induced level-set estimate. But from the theory, an explicit relation between $\gamma$ and $t_\gamma$ is hard to derive. By default we set $\gamma$ to $1$.
%
One could object that in some situations, it is useful to randomize this parameter. For instance, in the case of a bi-modal distribution for the inlier/normal behavior, one split of the tree needs to separate two clusters, in order for the level set estimate to distinguish between the two modes. As illustrated in \Cref{ocrf:fig:split_alpha_2}, it can only occur if $n_t'$ is large with respect to $n_t$ ($\gamma \gg 1$). However, the randomization of $\gamma$ is somehow included in the randomization of each tree, thanks to the sub-sampling inherent to RFs.
Moreover, small clusters tend to vanish when the sub-sample size is sufficiently small: a small sub-sampling size is used in \cite{Liu2008} to isolate outliers even when they form clusters.

\subsection{Alternative scoring functions.}\label{supp:scoring_functions}
Although we use the scoring function defined in \Cref{ocrf:eq:scoring3} because of its established high performance \cite{Liu2008}, other scoring functions can be defined.
A natural idea to adapt the majority vote to the one-class setting is to change the single vote of a leaf node $t$ into the fraction $\frac{n_t}{\leb(\mathcal{X}_t)}$, the forest output being the average of the latter quantity over the forest,
$s(x) = \sum_{t \text{~leaves}} \mathds{1}_{\{ x \in t \}} \frac{n_t}{\leb(\mathcal{X}_t)}.$
In such a case, each tree of the forest yields a piece-wise density estimate on its induced partition. 
The output produced by the forest is then a \emph{step-wise density estimate}.
%
We could also think about the \emph{local density of a typical cell}.
For each point $x$ of the input space, it returns the average number of observations in the leaves containing $x$, divided by the average volume of such leaves.
The output of OneClassRF is then the scoring function
$s(x) = \big(\sum_{t \text{~leaves}} \mathds{1}_{\{ x \in t \}} n_t\big) \big(\sum_{t \text{~leaves}} \mathds{1}_{\{ x \in t \}} \leb(\mathcal{X}_t)\big)^{-1}$, where the sums are over each leave of each tree in the forest.
This score can be interpreted as the local density of a `typical' cell (typical among those usually containing $x$).
%

\subsection{Alternative stopping criteria}\label{supp:stopping_criteria}
Other stopping criteria than a maximal depth may be considered. We could stop splitting a node $t$ when it contains less than $n\_min$ observations, or when the quantity $n_t/\leb(\mathcal{X}_t)$ is large enough (all the points in the cell $\mathcal{X}_t$ are likely to be inliers) or close enough to $0$ (all the points in the cell $\mathcal{X}_t$ are likely to be outliers). These options are not discussed in this work.

\subsection{Variable importance}
In the multi-class setting, \cite{Breiman2001} proposed to evaluate the importance of a feature $j \in \{1,\ldots d\}$ for prediction by 
 adding up the weighted impurity decreases 
for all nodes $t$ where $X_j$ is used, averaged over all the trees. The analogue quantity can be computed with respect to the one-class impurity decrease proxy. 
In our one-class setting, this quantity represents the size of the tail of $X_j$, and can be interpreted as the capacity of feature $j$ to discriminate between inliers/outliers.

\pgfmathsetseed{7}

\section{Hyper-parameters of tested algorithms}\label{supp:hyper_choice}

Overall we chose to train the different algorithms with their (default) hyper-parameters as seen
in their respective paper or author's implementation.
Indeed, since we are in an unsupervised setting, there is no trivial way to select/learn the hyperparameters of the different algorithm in the training phase
-- the labels are not supposed to be available. Hence the more realistic way to test the algorithms is to use their recommended/default hyperparameters.

The \emph{OCSVM} algorithm uses default parameters: \verb+kernel='rbf'+, \verb+tol=1e-3+,
\verb+nu=0.5+, \verb+shrinking=True+, \verb+gamma=1/n_features+, where tol is the tolerance for stopping criterion.

The \emph{LOF} algorithm uses default parameters:
\verb+n_neighbors=5+, \verb+leaf_size=30+, \verb+metric='minkowski'+,
\verb+contamination=0.1+, \verb+algorithm='auto'+, where the algorithm parameters stipulates
how to compute the nearest neighbors (either ball-tree, kd-tree or brute-force).

The \emph{iForest} algorithm uses default parameters:
\verb+n_estimators=100+, \verb+max_samples=min(256, n_samples)+,
\verb+max_features=1+, \verb+bootstrap=false+, where bootstrap states whether samples are drawn with replacement.

The \emph{OCRFsampling} algorithm uses default parameters:
the number of dimensions for the Random Subspace Method \verb+krsm=-1+,
the number of features randomly selected at each node during the induction of the tree \verb+krfs=-1+,
\verb+n_tree=100+,
the factor controlling the extension of the outlier domain used to sample outliers according to the volume of the hyper-box surrounding the target data \verb+alpha=1.2+,
the factor controlling the number of outlier data generated according to the number of target data \verb+beta=10+,
whether outliers are generated from uniform distribution \verb+optimize=0+,
whether data outside target bounds are considered as outlier data \verb+rejectOutOfBounds=0+.

The \emph{Orca} algorithm uses default parameter \verb+k=5+ (number of nearest neighbors)
as well as \verb+N=n/8+ (how many anomalies are to be reported).
The last setting, set up in the empirical evaluation of iForest in \cite{Liu2012},
allows a better computation time without impacting Orca's performance.

The \emph{RFC} algorithm uses default parameters:
\verb+no.forests=25+, \verb+no.trees=3000+,
the Addcl1 Random Forest dissimilarity \verb+addcl1=T, addcl2=F+,
use the importance measure \verb+imp=T+,
the data generating process \verb+oob.prox1=T+,
the number of features sampled at each split \verb+mtry1=3+.

The \emph{LSAD} algorithm uses default parameters:
the maximum number of samples per kernel \verb+n_kernels_max=500+,
the center of each kernel (the center of the random sample subset by default) \verb+kernel_pos='None'+,
the kernel scale parameter (using the pairwise median trick by default)\verb+gamma='None'+,
the regularization parameter \verb+rho=0.1+.

\section{Description of the datasets}\label{supp:dataset_description}

The characteristics of the twelve reference datasets considered here are summarized
in \Cref{ocrf:table:data}. They are all available on the UCI repository
\cite{Lichman2013} and the preprocessing is done in a classical way. 
In anomaly detection, we typically have data from two class (inliers/outliers) -- in novelty detection, the second class is unavailable in training in outlier detection, training data are polluted by second class (anonymous) examples. The classical approach to adapt multi-class data to this framework is to set classes forming the outlier class, while the other classes form the inlier class.

We removed all categorial attributes. Indeed, our method is designed to handle data whose distribution is absolutely continuous \wrt~the Lebesgue measure. 
The \emph{http} and \emph{smtp} datasets belong to the KDD Cup '99 dataset \cite{KDD99, Tavallaee2009}, which consist of a wide variety of hand-injected  attacks (anomalies) in a closed network (normal/inlier background). They are classically obtained as described in \cite{Yamanishi2000}. This two datasets are available on the \emph{scikit-learn} library \cite{sklearn2011}.
The \emph{shuttle} dataset is the fusion of the training and testing datasets available in the UCI repository. As in \cite{Liu2008}, we use instances from all different classes but class $4$. 
In the \emph{forestcover} data, the inliers are the instances from class~$2$ while instances from class $4$ are anomalies (as in \cite{Liu2008}). 
The \emph{ionosphere} dataset differentiates `good' from `bad' radars, considered here as abnormal. A `good' radar shows evidence of some type of structure in the ionosphere. A `bad' radar does not, its signal passing through the ionosphere.
The \emph{spambase} dataset consists of spam or non-spam emails. The former constitute our anomaly class.
The \emph{annthyroid} medical dataset on hypothyroidism contains one normal class and two abnormal ones, which form our outliers.
The \emph{arrhythmia} dataset reflects the presence and absence (class $1$) of cardiac arrhythmia. The number of attributes being large considering the sample size, we removed attributes containing missing data.
Besides, we removed attributes taking less than $10$ different values, the latter breaking too strongly our absolutely continuous assumption (\wrt~to $\leb$).
The \emph{pendigits} dataset contains 10 classes corresponding to the digits from 0 to 9, examples being handwriting samples. As in \cite{Schubert2012}, the outliers are chosen to be those from class 4.
The \emph{pima} dataset consists of medical data on diabetes. Patients suffering from diabetes (inlier class) were considered outliers.
The \emph{wild} dataset involves detecting diseased trees in Quickbird imagery. Diseased trees (class `w') is our outlier class.
In the \emph{adult} dataset, the goal is to predict whether income exceeds \$ 50K/year based on census data. We only keep the 6 continuous attributes.

\section{Further details on benchmarks and outlier detection results}
\label{sup:outlier_detection}

\Cref{ocrf:figresultssemisupervised} shows that the amount of time
to train\protect\footnotemark and test any dataset takes less than one minute with OneClassRF,
whereas some algorithms have far higher computation times (OCRFsampling,
OneClassSVM, LOF and Orca have computation times higher than 30 minutes in some
datasets). Our approach yields results similar to quite new algorithms such as
iForest and LSDA.
\footnotetext{\small For OCRF, Orca and RFC, testing and training time cannot be isolated
because of algorithms implementation: for these algorithms, the sum of the training and testing times are displayed in \Cref{ocrf:figresultssemisupervised} and \Cref{ocrf:figresultsunsupervised}.}

\pgfplotsset{minor grid style={very thick,black}}
\begin{figure}[ht]
\centering
\definecolor{ggreen}{rgb}{0.3,0.7,0.4}
\definecolor{ggreen2}{rgb}{0.4,0.8,0.5}
\definecolor{orange2}{rgb}{1,0.7,0}
\definecolor{violette}{rgb}{0.7,0.15,0.9}
\begin{tikzpicture}[scale=0.6,font=\Large]
\begin{axis}[ at={(0,0)},
              grid=minor,
              width=23cm, height=6cm,
              ybar=0pt,
              minor xtick={0.5,1.5,...,12.5},
              xmin=0.5, xmax=12.5,
              xticklabels={ , , , , , , , , , , , },
              ymin=0.58,
              ytick={0.6,0.7,...,1},
              ymax=1,
              ylabel={ROC AUC},
              legend entries={OneClassRF~~~~,iForest~~~~, OCRFsampling~~~~, OneClassSVM~~~~, LOF~~~~, Orca~~~~, LSAD~~~~, RFC},
legend style={at={(0.5,1.16)}, anchor=north,legend columns=-1}
              ]
\draw[dashed,black!30] (axis cs:0,0.6)--(axis cs:12.5,0.6);
\draw[dashed,black!30] (axis cs:0,0.7)--(axis cs:12.5,0.7);
\draw[dashed,black!30] (axis cs:0,0.8)--(axis cs:12.5,0.8);
\draw[dashed,black!30] (axis cs:0,0.9)--(axis cs:12.5,0.9);
\addplot+[bar width=4.5pt] plot table[x index=1, y index=3]{results_semisupervised.txt};
\addplot+[bar width=4.5pt] plot table[x index=1, y index=7]{results_semisupervised.txt};
\addplot+[bar width=4.5pt] plot table[x index=1, y index=11]{results_semisupervised.txt};
\addplot+[bar width=4.5pt] plot table[x index=1, y index=15]{results_semisupervised.txt};
\addplot+[bar width=4.5pt, fill=ggreen!50, draw=ggreen] plot table[x index=1, y index=19]{results_semisupervised.txt};
\addplot+[bar width=4.5pt, fill=violette!50, draw=violette] plot table[x index=1, y index=23]{results_semisupervised.txt};
\addplot+[bar width=4.5pt, fill=orange2!50, draw=orange2] plot table[x index=1, y index=27]{results_semisupervised.txt};
\addplot+[bar width=4.5pt,fill=white, draw=black!40] plot table[x index=1, y index=31]{results_semisupervised.txt};
\end{axis}
\begin{axis}[ at={(0,-5cm)},
              grid=minor,
              width=23cm, height=6cm,
              ybar=0pt,
              minor xtick={0.5,1.5,...,12.5},
              xmin=0.5, xmax=12.5,
              xticklabels={ , , , , , , , , , , , },
              ymin=0.0,
              ytick={0.2,0.4,0.6,0.8,1},
              ymax=1,
              ylabel={PR AUC},
legend style={at={(0.5,1.16)}, anchor=north,legend columns=-1}
              ]
\draw[dashed,black!30] (axis cs:0,0.2)--(axis cs:12.5,0.2);
\draw[dashed,black!30] (axis cs:0,0.4)--(axis cs:12.5,0.4);
\draw[dashed,black!30] (axis cs:0,0.6)--(axis cs:12.5,0.6);
\draw[dashed,black!30] (axis cs:0,0.8)--(axis cs:12.5,0.8);
\addplot+[bar width=4.5pt] plot table[x index=1, y index=5]{results_semisupervised.txt};
\addplot+[bar width=4.5pt] plot table[x index=1, y index=9]{results_semisupervised.txt};
\addplot+[bar width=4.5pt] plot table[x index=1, y index=13]{results_semisupervised.txt};
\addplot+[bar width=4.5pt] plot table[x index=1, y index=17]{results_semisupervised.txt};
\addplot+[bar width=4.5pt, fill=ggreen!50, draw=ggreen] plot table[x index=1, y index=21]{results_semisupervised.txt};
\addplot+[bar width=4.5pt, fill=violette!50, draw=violette] plot table[x index=1, y index=25]{results_unsupervised.txt};
\addplot+[bar width=4.5pt, fill=orange2!50, draw=orange2] plot table[x index=1, y index=29]{results_semisupervised.txt};
\addplot+[bar width=4.5pt,fill=white, draw=black!40] plot table[x index=1, y index=33]{results_semisupervised.txt};
\end{axis}
\begin{axis}[  at={(0,-13.4)},
              grid=minor,
              width=23cm,height=6cm,
              ybar=0pt,
              minor xtick={0.5,1.5,...,12.5},
              xmin=0.5, xmax=12.5,
              xtick={1,...,12},
              xticklabels={adult, annthyroid, arrhythmia, forestcover, http, ionosphere, pendigits, pima, shuttle, smtp, spambase, wilt},
              x tick label style={rotate=20,anchor=east},
              ymax=60, ymin=0,
              ytick={0,10,...,60},
              ylabel={Computation time (sec.)}
              ]
\draw[dashed,black!30] (axis cs:0,10)--(axis cs:12.5,10);
\draw[dashed,black!30] (axis cs:0,20)--(axis cs:12.5,20);
\draw[dashed,black!30] (axis cs:0,30)--(axis cs:12.5,30);
\draw[dashed,black!30] (axis cs:0,40)--(axis cs:12.5,40);
\draw[dashed,black!30] (axis cs:0,50)--(axis cs:12.5,50);
\addplot+[bar width=4.5pt] plot table[x index=1, y index=2]{computationTime_semisupervised.txt};
\addplot+[bar width=4.5pt] plot table[x index=1, y index=5]{computationTime_semisupervised.txt};
\addplot+[bar width=4.5pt] plot table[x index=1, y index=8]{computationTime_semisupervised.txt};
\addplot+[bar width=4.5pt] plot table[x index=1, y index=9]{computationTime_semisupervised.txt};
\addplot+[bar width=4.5pt, fill=ggreen!50, draw=ggreen] plot table[x index=1, y index=12]{computationTime_semisupervised.txt};
\addplot+[bar width=4.5pt, fill=violette!50, draw=violette] plot table[x index=1, y index=15]{computationTime_semisupervised.txt};
\addplot+[bar width=4.5pt, fill=orange2!50, draw=orange2] plot table[x index=1, y index=16]{computationTime_semisupervised.txt};
\addplot+[bar width=4.5pt,fill=white, draw=black!40] plot table[x index=1, y index=19]{computationTime_semisupervised.txt};
\end{axis}
\end{tikzpicture}
\caption{Performances of the algorithms on each dataset in the novelty detection framework:
ROC AUCs are displayed on the top, Precision-Recall AUCs in the middle and training times on the bottom,
for each dataset and algorithm. The $x$-axis represents the datasets.}
\label{ocrf:figresultssemisupervised}
\end{figure}

In this section present experiments in the outlier detections setting. For each algorithm, 10 experiments on random training and testing datasets are performed. Averaged ROC and Precision-Recall curves AUC are summarized in \Cref{ocrf:table:results-unsupervised}.
For the experiments made in an unsupervised framework (meaning that the training set is polluted by outliers), the anomaly rate is arbitrarily bounded to $10\%$ max (before splitting data into training and testing sets).

\begin{table*}[h]
\caption{Results for the outlier detection setting}
\label{ocrf:table:results-unsupervised}

\centering
\tabcolsep=0.1cm
\resizebox{\linewidth}{!} {
\begin{tabular}{ l  c@{\extracolsep{0.1cm}}c c@{\extracolsep{0.1cm}}c c@{\extracolsep{0.1cm}}c c@{\extracolsep{0.1cm}}c c@{\extracolsep{0.1cm}}c c@{\extracolsep{0.1cm}}c c@{\extracolsep{0.1cm}}c c@{\extracolsep{0.1cm}}c }
\toprule

Dataset & \multicolumn{2}{c }{OneClassRF} & \multicolumn{2}{c }{iForest} & \multicolumn{2}{c }{OCRFsampling} & \multicolumn{2}{c }{OCSVM}& \multicolumn{2}{c }{LOF}& \multicolumn{2}{c }{Orca}& \multicolumn{2}{c }{LSDA}& \multicolumn{2}{c }{RFC}  \\
\cmidrule{1-17}
~            & ROC &  PR & ROC &  PR & ROC & PR  & ROC & PR  & ROC & PR  &ROC  & PR  & ROC &  PR & ROC & PR  \\
adult                 & 0.625 & 0.161   & \textbf{0.644} & 0.234   & NA   & NA     & 0.622 & 0.179   & 0.546 & 0.100   & 0.593 & 0.179   & 0.633 & 0.204       & NA   & NA  \\
annthyroid             & 0.842 & 0.226   & 0.820 & 0.310   & \textbf{0.992} & 0.869   & 0.688 & 0.193   & 0.731 & 0.188   & 0.561 & 0.132   & 0.762 & 0.246       & NA   & NA  \\
arrhythmia             & 0.698 & 0.485   & 0.746 & 0.418   & 0.704 & 0.276   & \textbf{0.916} & 0.630   & 0.765 & 0.468   & 0.741 & 0.502   & 0.733 & 0.393       & 0.711 & 0.309 \\
forestcover             & 0.845 & 0.044   & \textbf{0.882} & 0.062   & NA   & NA     & NA & NA   & 0.550 & 0.017   & 0.696 & 0.045   & 0.816 & 0.072       & NA   & NA   \\
http                  & 0.984 & 0.120   & \textbf{0.999} & 0.685   & NA   & NA    & NA & NA   & NA   & NA     & 0.998 & 0.402   & 0.277 & 0.074       & NA   & NA   \\
ionosphere              & 0.903 & 0.508   & 0.888 & 0.545   & 0.879 & 0.664   & \textbf{0.956} & 0.813   & \textbf{0.956} & 0.789   & 0.929 & 0.917   & 0.915 & 0.773       & 0.943 & 0.725 \\
pendigits             & 0.453 & 0.085   & 0.463 & 0.077   & \textbf{0.999} & 0.993   & 0.366 & 0.066   & 0.491 & 0.086   & 0.495 & 0.086   & 0.513 & 0.091       & NA   & NA   \\
pima                  & 0.708 & 0.229   & 0.743 & 0.205   & \textbf{0.790} & 0.296   & 0.706 & 0.226   & 0.670 & 0.137   & 0.585 & 0.170   & 0.686 & 0.190       & 0.505 & 0.091 \\
shuttle               & 0.947 & 0.491   & \textbf{0.997} & 0.979   & NA   & NA     & 0.992 & 0.904   & 0.526 & 0.115   & 0.655 & 0.320   & 0.686 & 0.218       & NA  & NA   \\
smtp                  & \textbf{0.916} & 0.400   & 0.902 & 0.005   & NA   & NA     & 0.881 & 0.372   & 0.909 & 0.053   & 0.824 & 0.236   & 0.888 & 0.398       & NA   & NA   \\
spambase             & 0.830 & 0.300   & 0.799 & 0.303   & \textbf{0.970} & 0.877   & 0.722 & 0.192   & 0.664 & 0.120   & 0.603 & 0.210   & 0.731 & 0.229       & 0.684 & 0.134 \\
wilt                 & 0.520 & 0.053   & 0.443 & 0.044   & \textbf{0.966} & 0.554   & 0.316 & 0.036   & 0.627 & 0.069   & 0.441 & 0.029   & 0.530 & 0.053       & 0.876 & 0.472 \\
\cmidrule{1-17}
average: & 0.773 &  0.259 & 0.777 & 0.322 & \textbf{0.900} & 0.647 & 0.717 & 0.361 & 0.676 & 0.195 & 0.677 & 0.269 & 0.681 & 0.245 & 0.744 & 0.346 \\
cum. train time: & \multicolumn{2}{c }{\textbf{61s}} & \multicolumn{2}{c }{70s} & \multicolumn{2}{c }{NA} & \multicolumn{2}{c }{NA}& \multicolumn{2}{c }{NA}& \multicolumn{2}{c }{2432s}& \multicolumn{2}{c }{72s}& \multicolumn{2}{c }{NA}  \\
  \bottomrule
\end{tabular}
}
\end{table*}

\pgfplotsset{minor grid style={very thick,black}}
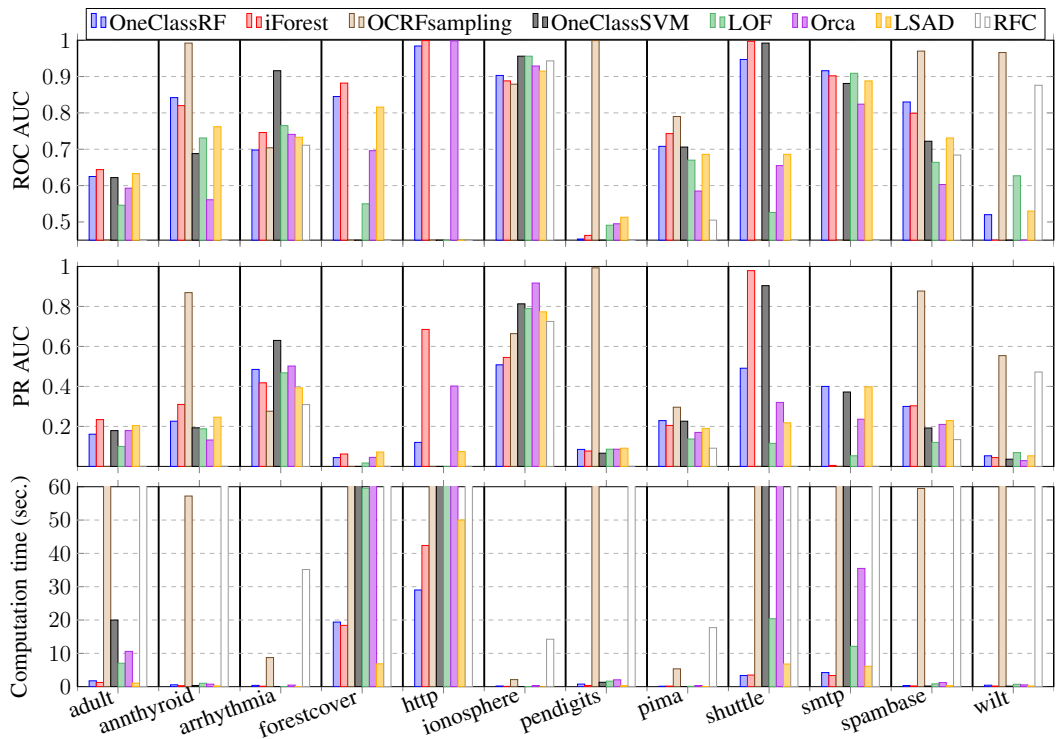
\begin{figure}
\centering
\definecolor{ggreen}{rgb}{0.3,0.7,0.4}
\definecolor{ggreen2}{rgb}{0.4,0.8,0.5}
\definecolor{orange2}{rgb}{1,0.7,0}
\definecolor{violette}{rgb}{0.7,0.15,0.9}
\begin{tikzpicture}[scale=0.6,font=\Large]
\begin{axis}[ at={(0,0)},
              grid=minor,
              width=23cm, height=6cm,
              ybar=0pt,
              minor xtick={0.5,1.5,...,12.5},
              xmin=0.5, xmax=12.5,
              xticklabels={ , , , , , , , , , , , },
              ymin=0.45,
              ytick={0.5,0.6,0.7,...,1},
              ymax=1,
              ylabel={ROC AUC},
              legend entries={OneClassRF~~~~,iForest~~~~, OCRFsampling~~~~, OneClassSVM~~~~, LOF~~~~, Orca~~~~, LSAD~~~~, RFC},
legend style={at={(0.5,1.16)}, anchor=north,legend columns=-1}
              ]
\draw[dashed,black!30] (axis cs:0,0.6)--(axis cs:12.5,0.6);
\draw[dashed,black!30] (axis cs:0,0.7)--(axis cs:12.5,0.7);
\draw[dashed,black!30] (axis cs:0,0.8)--(axis cs:12.5,0.8);
\draw[dashed,black!30] (axis cs:0,0.9)--(axis cs:12.5,0.9);
\addplot+[bar width=4.5pt] plot table[x index=1, y index=3]{results_unsupervised.txt};
\addplot+[bar width=4.5pt] plot table[x index=1, y index=7]{results_unsupervised.txt};
\addplot+[bar width=4.5pt] plot table[x index=1, y index=11]{results_unsupervised.txt};
\addplot+[bar width=4.5pt] plot table[x index=1, y index=15]{results_unsupervised.txt};
\addplot+[bar width=4.5pt, fill=ggreen!50, draw=ggreen] plot table[x index=1, y index=19]{results_unsupervised.txt};
\addplot+[bar width=4.5pt, fill=violette!50, draw=violette] plot table[x index=1, y index=23]{results_unsupervised.txt};
\addplot+[bar width=4.5pt, fill=orange2!50, draw=orange2] plot table[x index=1, y index=27]{results_unsupervised.txt};
\addplot+[bar width=4.5pt,fill=white, draw=black!40] plot table[x index=1, y index=31]{results_unsupervised.txt};
\end{axis}

\begin{axis}[  at={(0,-5cm)},
              grid=minor,
              width=23cm,height=6cm,
              ybar=0pt,
              minor xtick={0.5,1.5,...,12.5},
              xmin=0.5, xmax=12.5,
              xtick={1,...,12},
              xticklabels={ , , , , , , , , , , , },
              x tick label style={rotate=20,anchor=east},
              ytick={0.2,0.4,0.6,0.8,1},
              ymin=0,
              ymax=1,
              ylabel={PR AUC}
              ]
\draw[dashed,black!30] (axis cs:0,0.2)--(axis cs:12.5,0.2);
\draw[dashed,black!30] (axis cs:0,0.4)--(axis cs:12.5,0.4);
\draw[dashed,black!30] (axis cs:0,0.6)--(axis cs:12.5,0.6);
\draw[dashed,black!30] (axis cs:0,0.8)--(axis cs:12.5,0.8);
\addplot+[bar width=4.5pt] plot table[x index=1, y index=5]{results_unsupervised.txt};
\addplot+[bar width=4.5pt] plot table[x index=1, y index=9]{results_unsupervised.txt};
\addplot+[bar width=4.5pt] plot table[x index=1, y index=13]{results_unsupervised.txt};
\addplot+[bar width=4.5pt] plot table[x index=1, y index=17]{results_unsupervised.txt};
\addplot+[bar width=4.5pt, fill=ggreen!50, draw=ggreen] plot table[x index=1, y index=21]{results_unsupervised.txt};
\addplot+[bar width=4.5pt, fill=violette!50, draw=violette] plot table[x index=1, y index=25]{results_unsupervised.txt};
\addplot+[bar width=4.5pt, fill=orange2!50, draw=orange2] plot table[x index=1, y index=29]{results_unsupervised.txt};
\addplot+[bar width=4.5pt,fill=white, draw=black!40] plot table[x index=1, y index=33]{results_unsupervised.txt};
\end{axis}

\begin{axis}[  at={(0,-13.4)},
              grid=minor,
              width=23cm,height=6cm,
              ybar=0pt,
              minor xtick={0.5,1.5,...,12.5},
              xmin=0.5, xmax=12.5,
              xtick={1,...,12},
              xticklabels={adult, annthyroid, arrhythmia, forestcover, http, ionosphere, pendigits, pima, shuttle, smtp, spambase, wilt},
              x tick label style={rotate=20,anchor=east},
              ymax=60, ymin=0,
              ytick={0,10,...,60},
              ylabel={Computation time (sec.)}
              ]
\draw[dashed,black!30] (axis cs:0,10)--(axis cs:12.5,10);
\draw[dashed,black!30] (axis cs:0,20)--(axis cs:12.5,20);
\draw[dashed,black!30] (axis cs:0,30)--(axis cs:12.5,30);
\draw[dashed,black!30] (axis cs:0,40)--(axis cs:12.5,40);
\draw[dashed,black!30] (axis cs:0,50)--(axis cs:12.5,50);
\addplot+[bar width=4.5pt] plot table[x index=1, y index=2]{computationTime_unsupervised.txt};
\addplot+[bar width=4.5pt] plot table[x index=1, y index=5]{computationTime_unsupervised.txt};
\addplot+[bar width=4.5pt] plot table[x index=1, y index=8]{computationTime_unsupervised.txt};
\addplot+[bar width=4.5pt] plot table[x index=1, y index=9]{computationTime_unsupervised.txt};
\addplot+[bar width=4.5pt, fill=ggreen!50, draw=ggreen] plot table[x index=1, y index=12]{computationTime_unsupervised.txt};
\addplot+[bar width=4.5pt, fill=violette!50, draw=violette] plot table[x index=1, y index=15]{computationTime_unsupervised.txt};
\addplot+[bar width=4.5pt, fill=orange2!50, draw=orange2] plot table[x index=1, y index=16]{computationTime_unsupervised.txt};
\addplot+[bar width=4.5pt,fill=white, draw=black!40] plot table[x index=1, y index=19]{computationTime_unsupervised.txt};
\end{axis}
\end{tikzpicture}
\caption{Performances of the algorithms on each dataset in the outlier detection framework:
ROC AUCs are on the top, Precision-Recall AUCs in the middle
and processing times are displayed below (for each dataset and algorithm).  The $x$-axis represents the datasets.}
\label{ocrf:figresultsunsupervised}
\end{figure}

\clearpage
\begin{figure}[!ht]
  \caption{ROC and PR curves for OneClassRF (novelty detection framework)}
  \label{ocrf:fig:oneclassrf_roc_pr}
  \centering
  \includegraphics[trim=175 80 175 123, clip, width=0.85\linewidth]{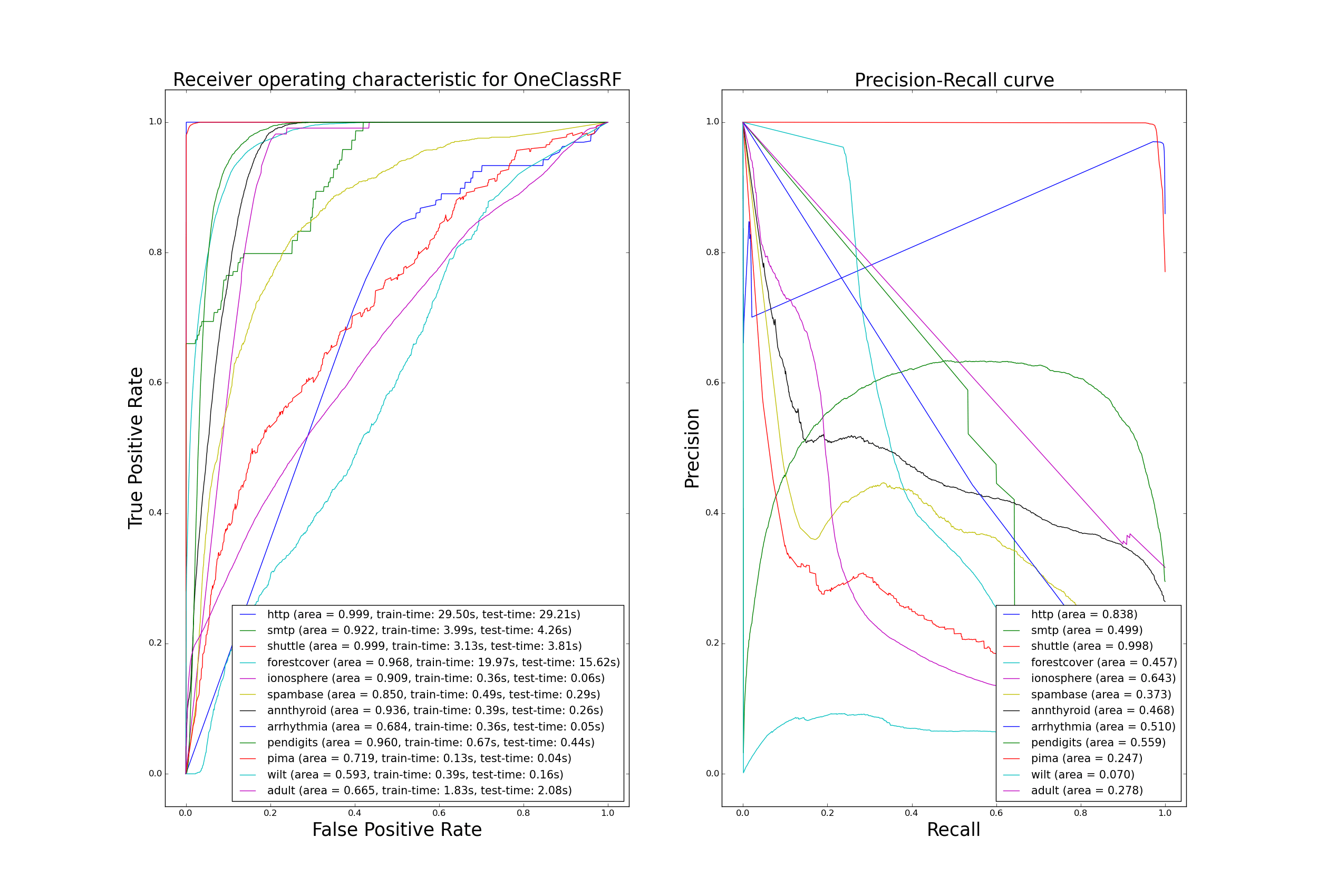}
\end{figure}
\begin{figure}[!ht]
  \caption{ROC and PR curves for OneClassRF (outlier detection framework)}
  \label{ocrf:fig:oneclassrf_roc_pr_unsupervised}
  \centering
  \includegraphics[trim=175 80 175 123, clip, width=0.85\linewidth]{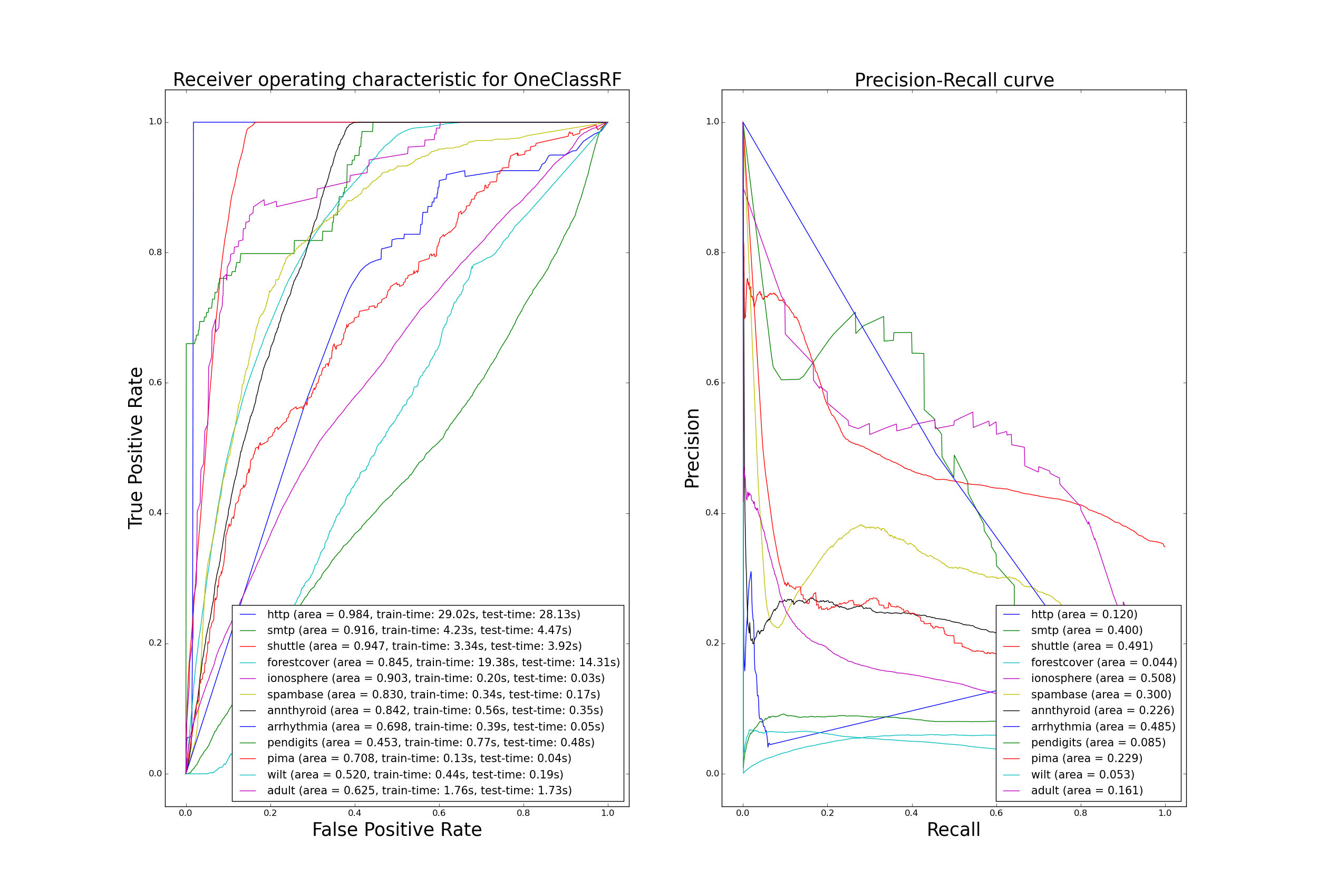}
\end{figure}

\begin{figure}[!ht]
  \caption{ROC and PR curves for IsolationForest (novelty detection framework)}
  \label{ocrf:fig:iforest_roc_pr}
  \centering
  \includegraphics[trim=175 80 175 123, clip, width=0.85\linewidth]{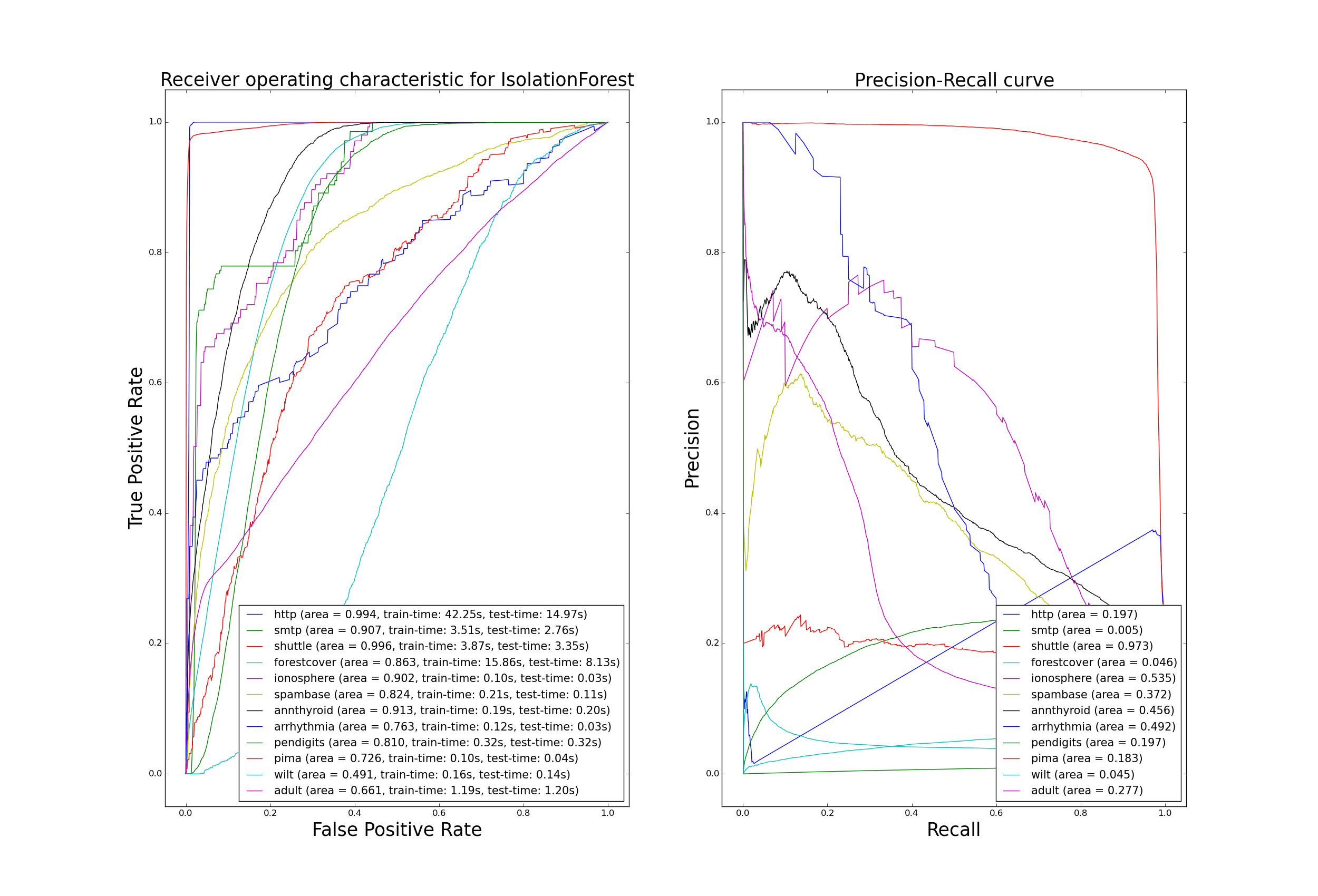}
\end{figure}
\begin{figure}[!ht]
  \caption{ROC and PR curves for IsolationForest (outlier detection framework)}
  \label{ocrf:fig:iforest_roc_pr_unsupervised}
  \centering
  \includegraphics[trim=175 80 175 123, clip, width=0.85\linewidth]{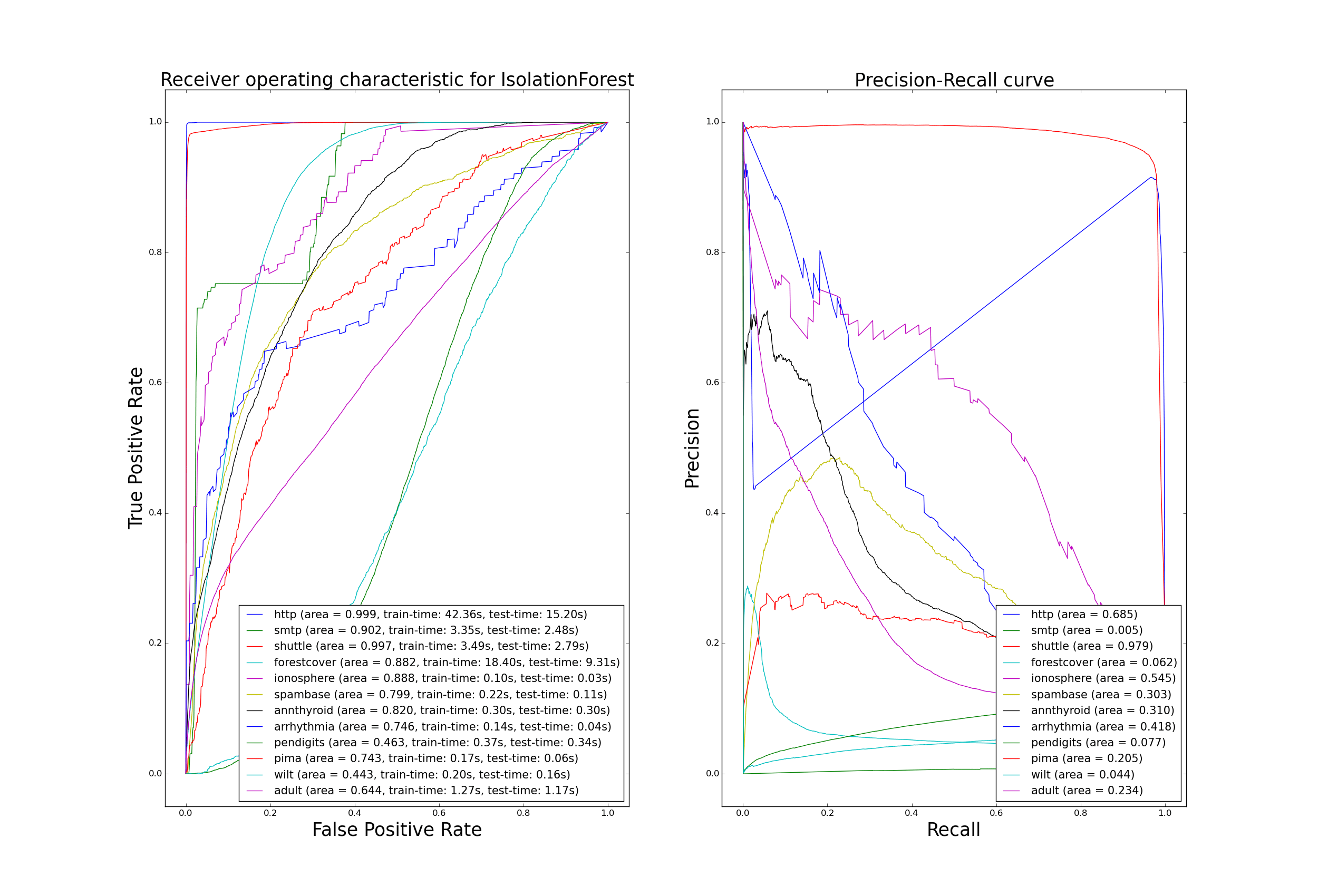}
\end{figure}

\begin{figure}[!ht]
  \caption{ROC and PR curves for OCRFsampling (novelty detection framework)}
  \label{ocrf:fig:ocrfm_roc_pr}
  \centering
  \includegraphics[trim=175 80 175 123, clip, width=0.85\linewidth]{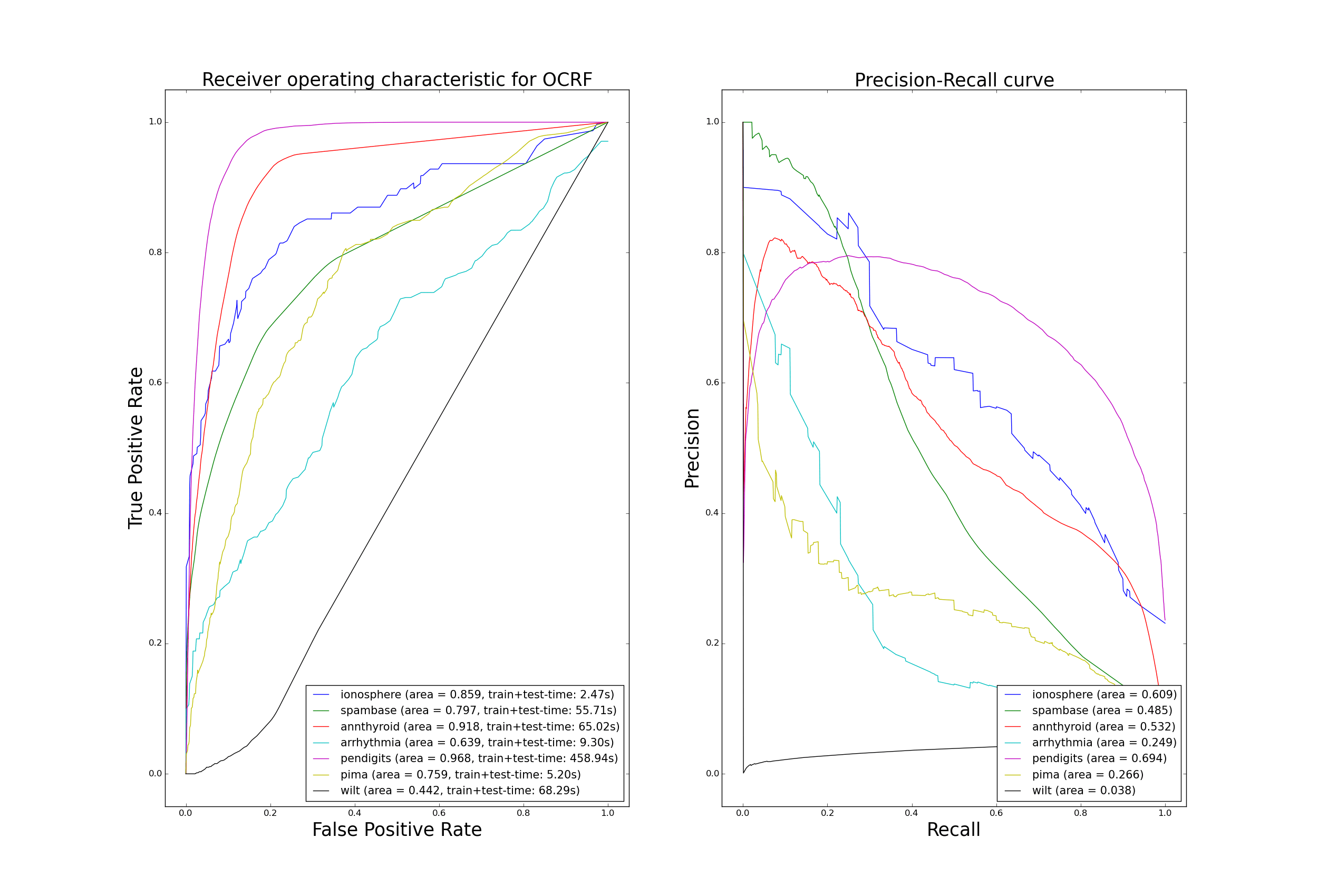}
\end{figure}
\begin{figure}[!ht]
  \caption{ROC and PR curves for OCRFsampling (outlier detection framework)}
  \label{ocrf:fig:ocrfm_roc_pr_unsupervised}
  \centering
  \includegraphics[trim=175 80 175 123, clip, width=0.85\linewidth]{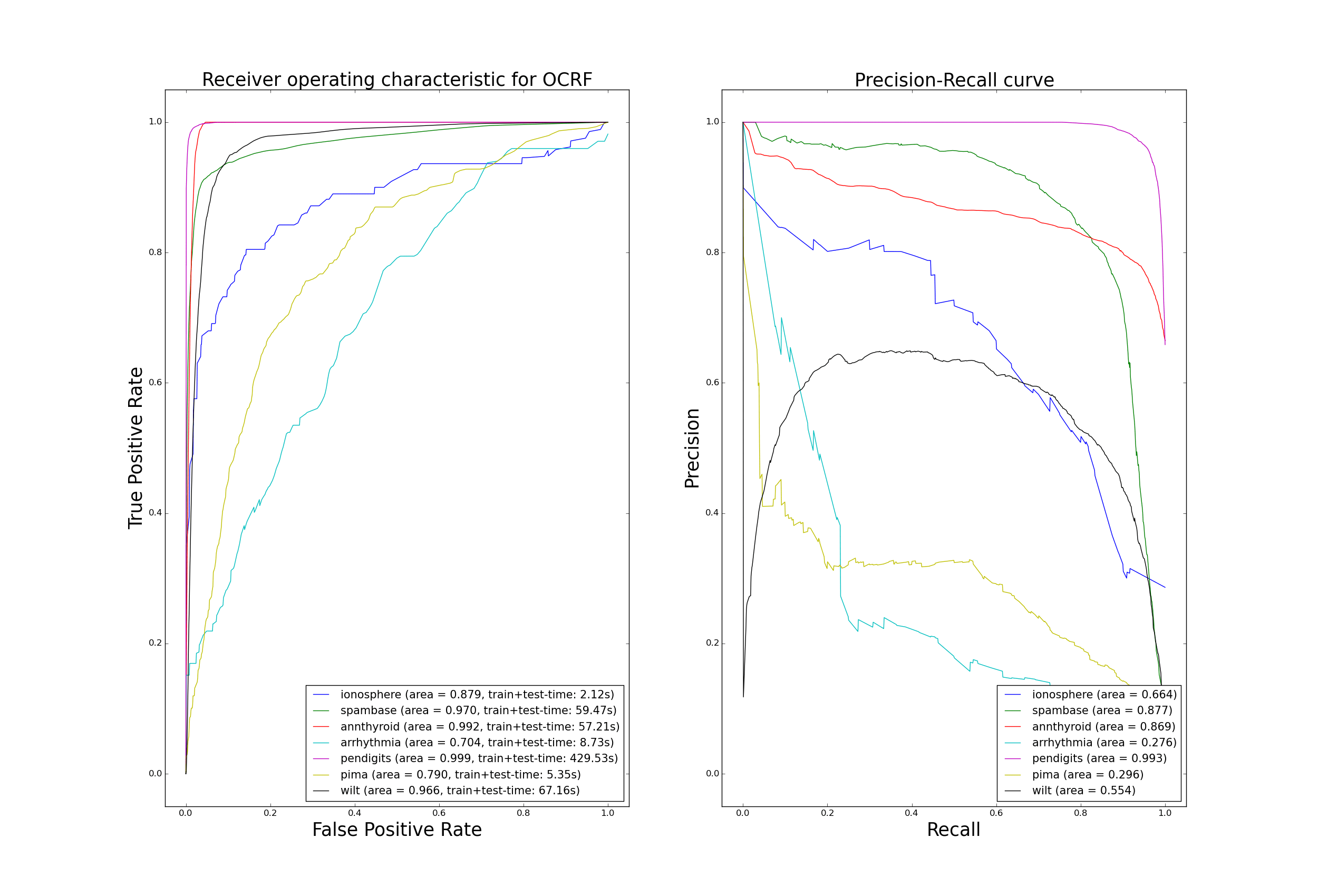}
\end{figure}

\begin{figure}[!ht]
  \caption{ROC and PR curves for OCSVM (novelty detection framework)}
  \label{ocrf:fig:ocsvm_roc_pr}
  \centering
  \includegraphics[trim=175 80 175 123, clip, width=0.85\linewidth]{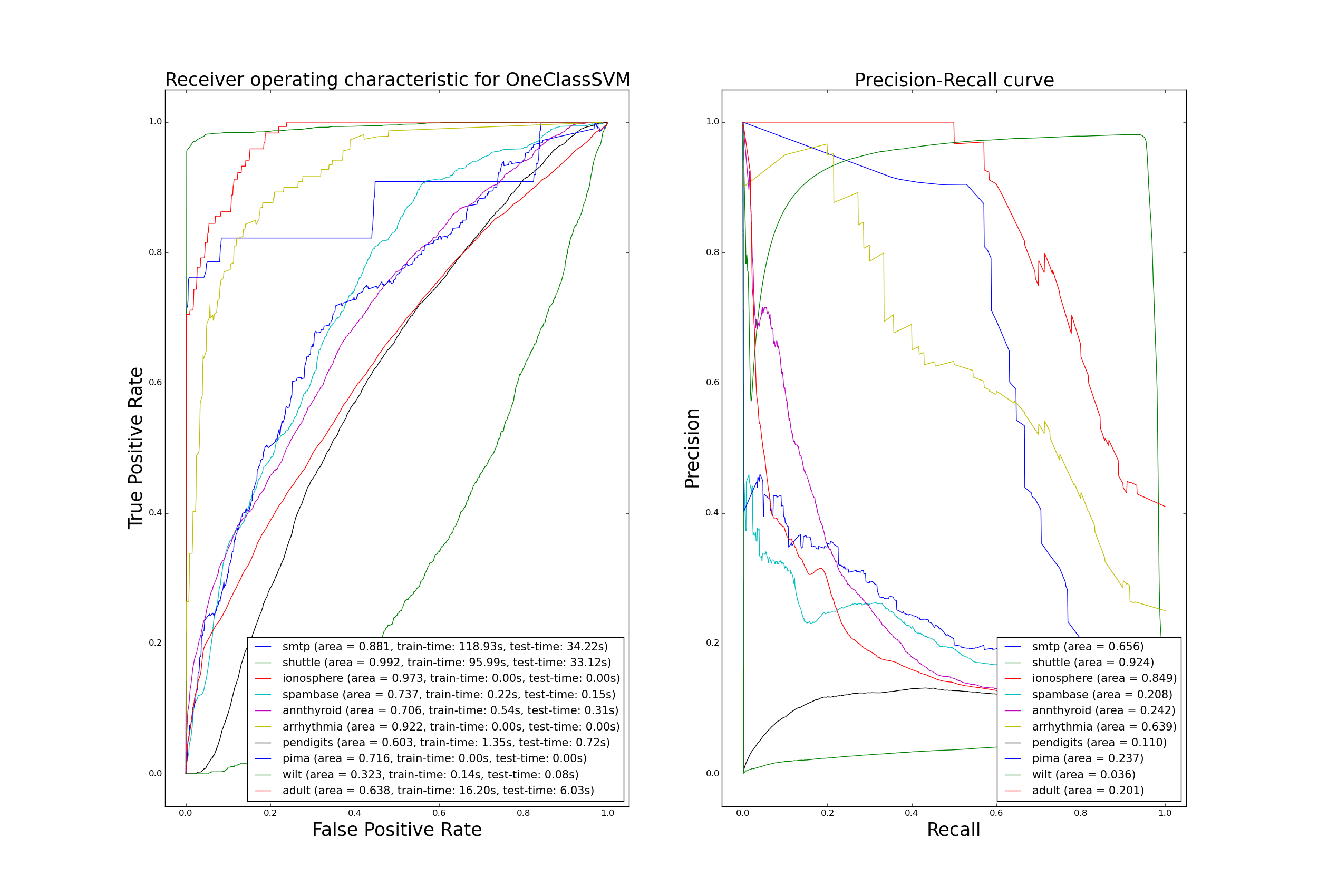}
\end{figure}
\begin{figure}[!ht]
  \caption{ROC and PR curves for OCSVM (outlier detection framework)}
  \label{ocrf:fig:ocsvm_roc_pr_unsupervised}
  \centering
  \includegraphics[trim=175 80 175 123, clip, width=0.85\linewidth]{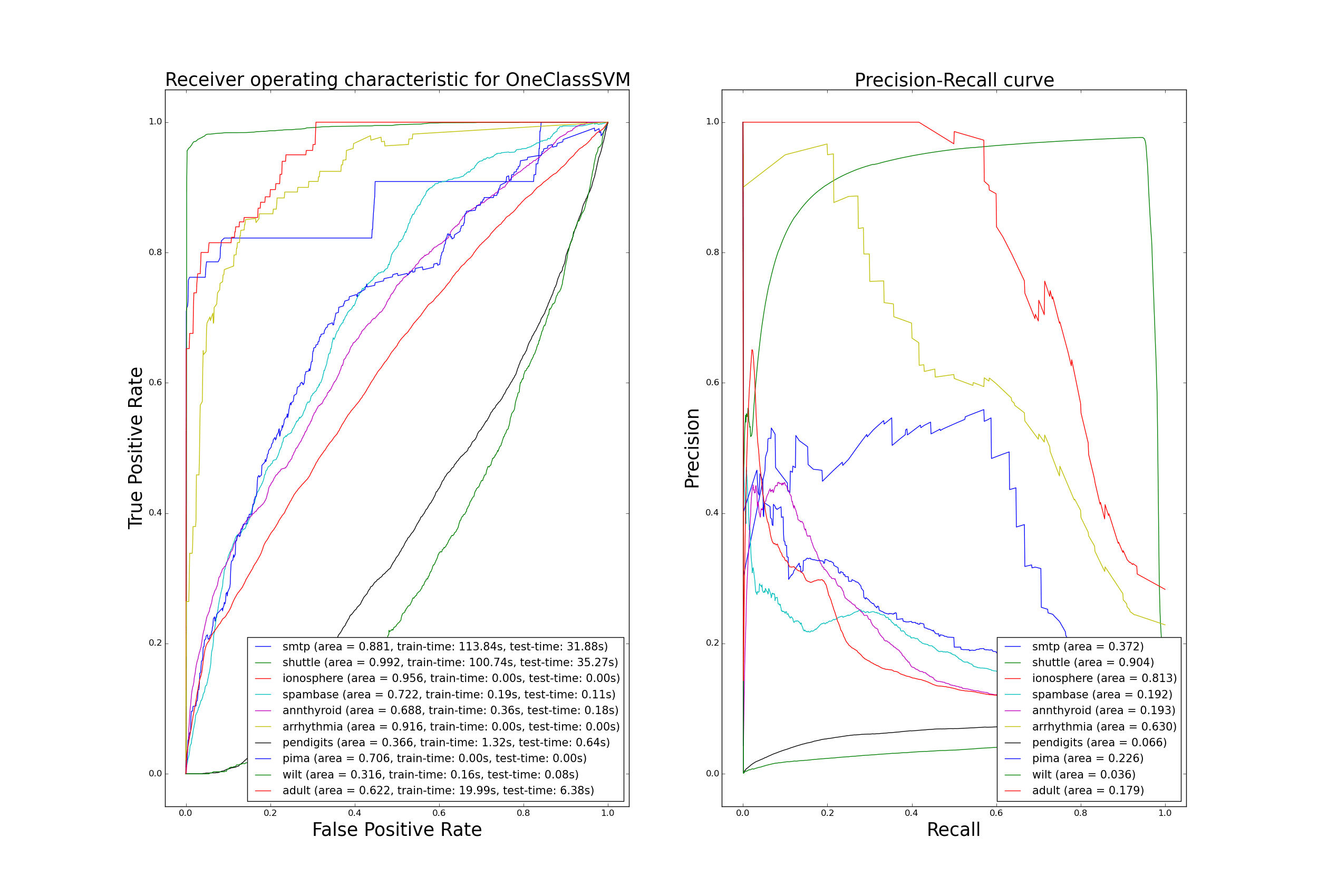}
\end{figure}

\begin{figure}[!ht]
  \caption{ROC and PR curves for LOF (novelty detection framework)}
  \label{ocrf:fig:lof_roc_pr}
  \centering
  \includegraphics[trim=175 80 175 123, clip, width=0.85\linewidth]{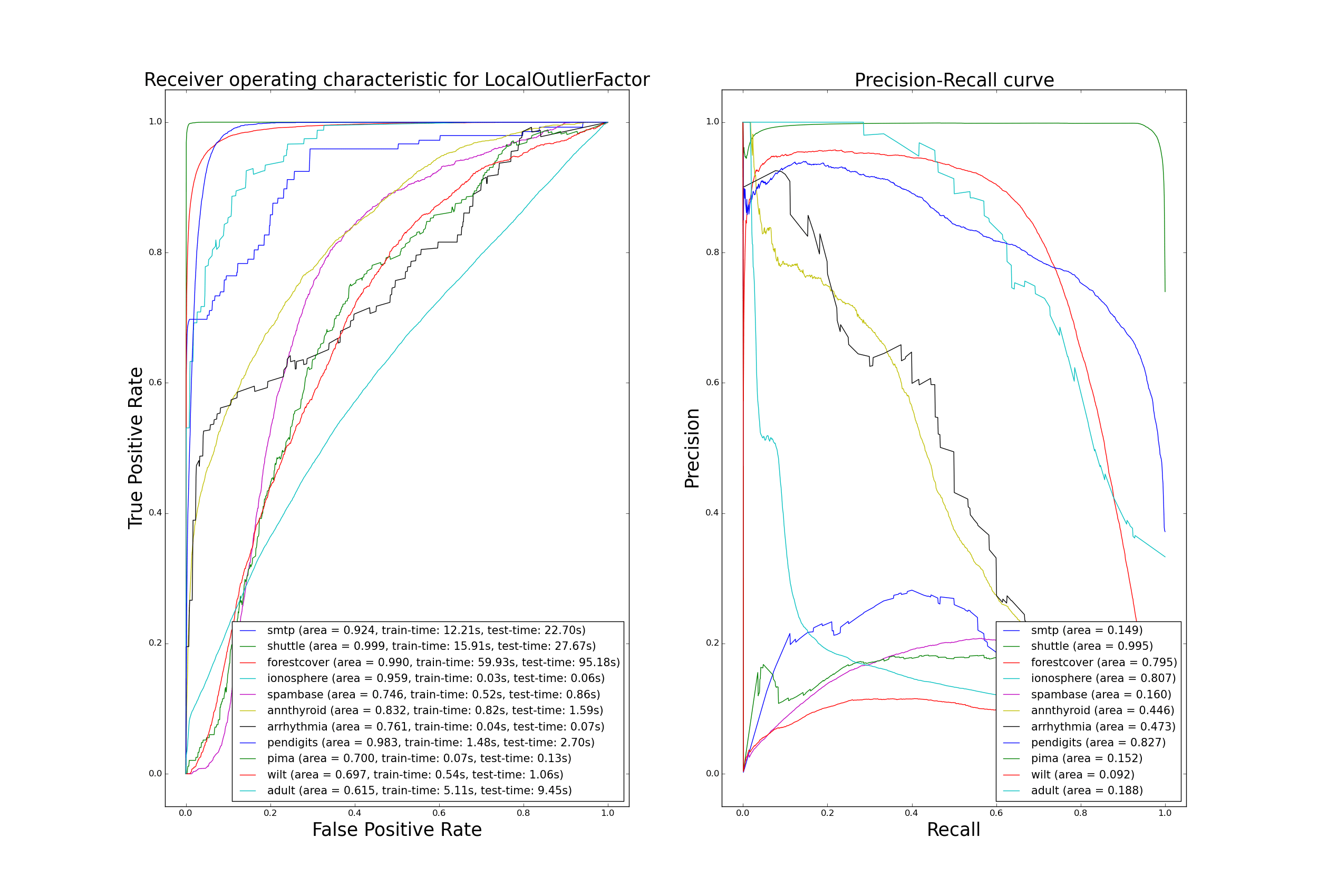}
\end{figure}
\begin{figure}[!ht]
  \caption{ROC and PR curves for LOF (outlier detection framework)}
  \label{ocrf:fig:lof_roc_pr_unsupervised}
  \centering
  \includegraphics[trim=175 80 175 123, clip, width=0.85\linewidth]{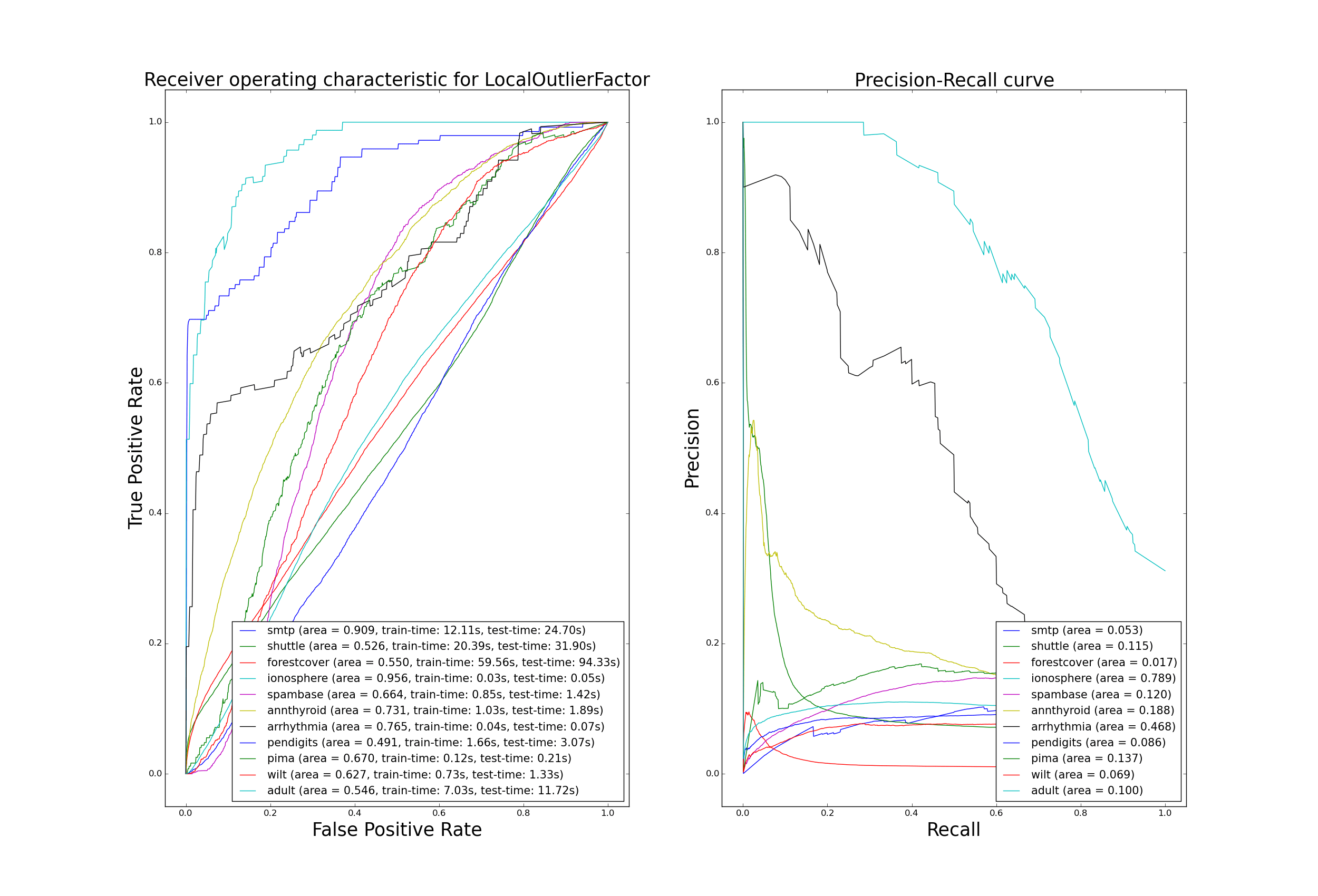}
\end{figure}

\begin{figure}[!ht]
  \caption{ROC and PR curves for Orca (novelty detection framework)}
  \label{ocrf:fig:orca_roc_pr}
  \centering
  \includegraphics[trim=175 80 175 123, clip, width=0.85\linewidth]{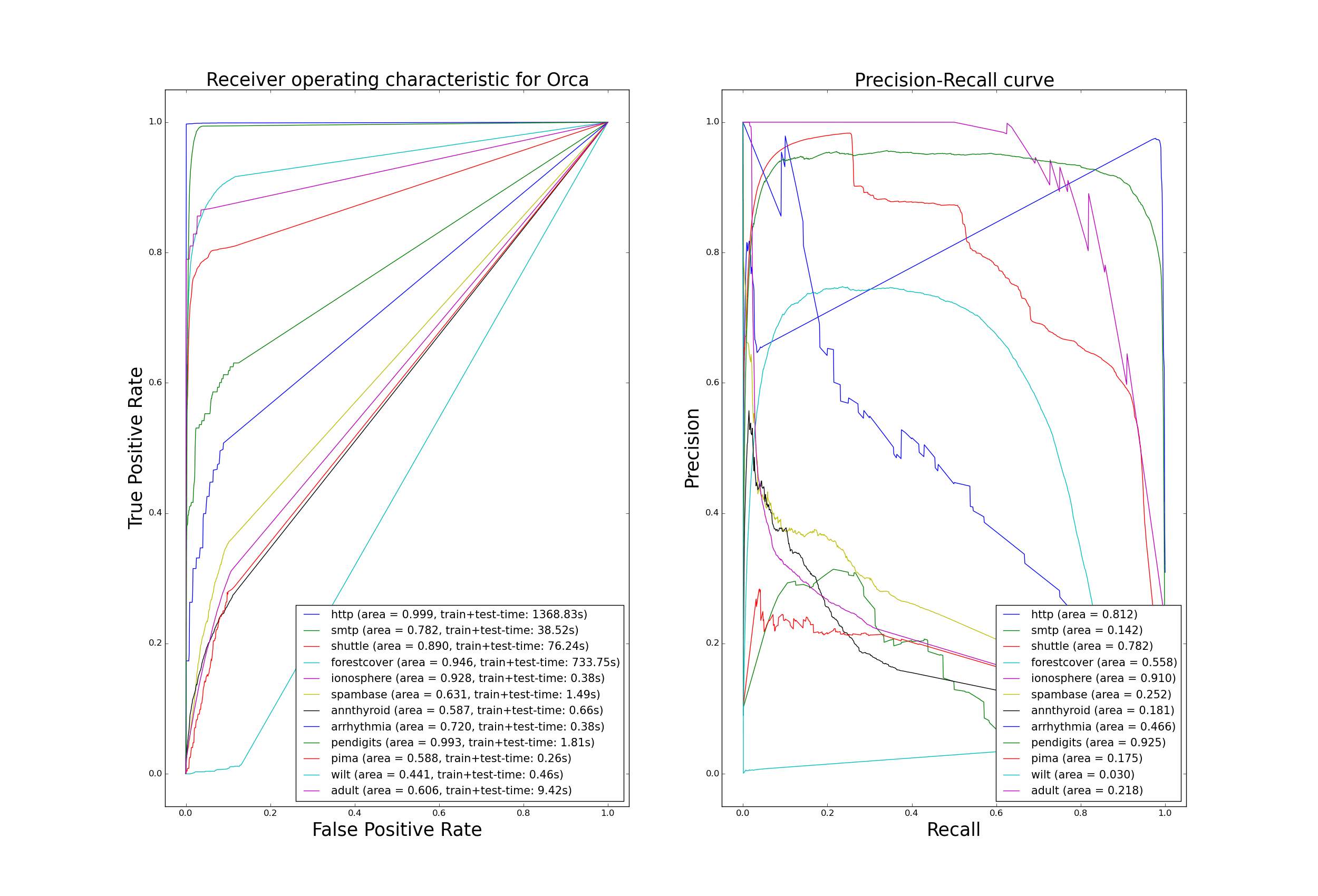}
\end{figure}
\begin{figure}[!ht]
  \caption{ROC and PR curves for Orca (outlier detection framework)}
  \label{ocrf:fig:orca_roc_pr_unsupervised}
  \centering
  \includegraphics[trim=175 80 175 123, clip, width=0.85\linewidth]{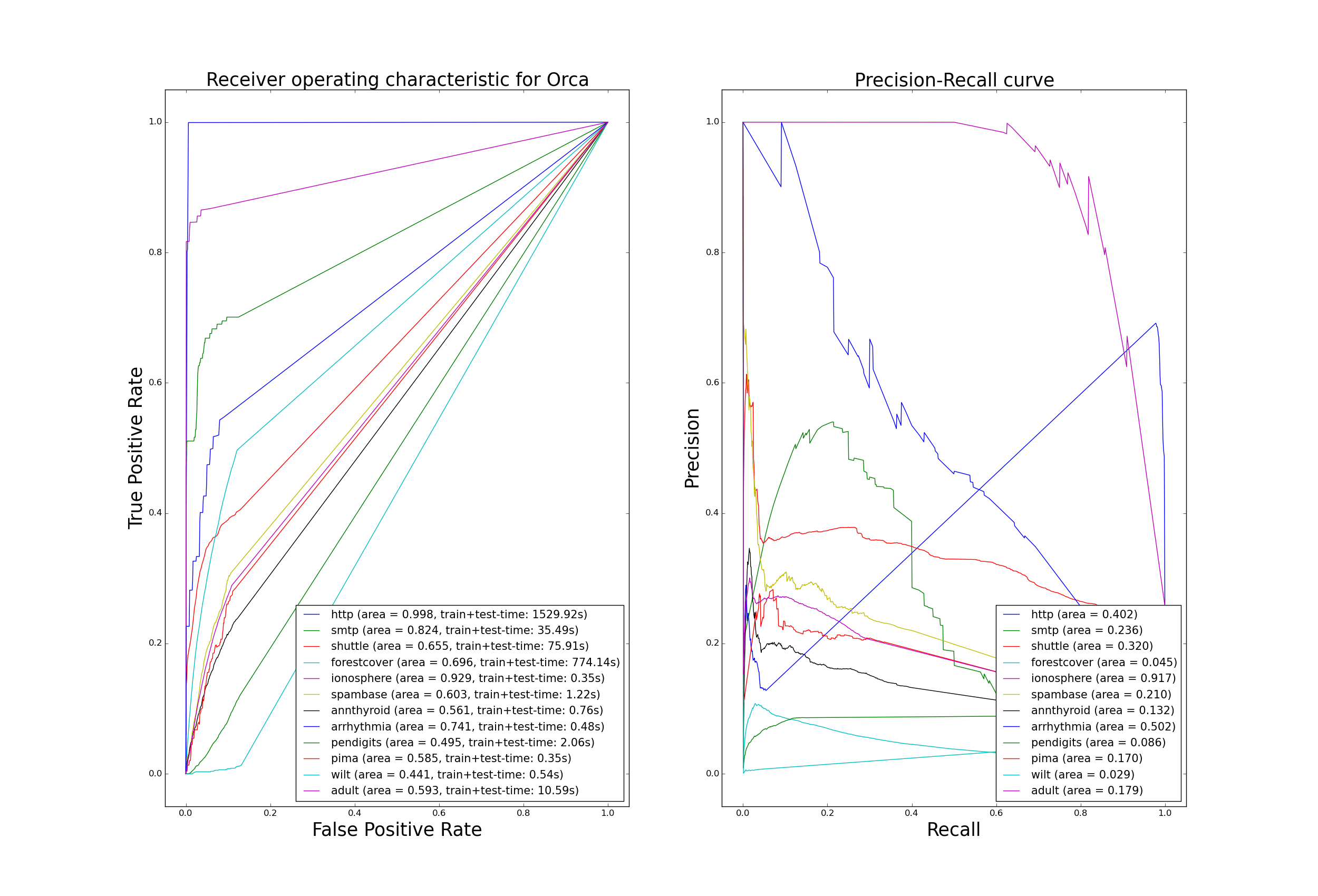}
\end{figure}

\begin{figure}[!ht]
  \caption{ROC and PR curves for LSAD (novelty detection framework)}
  \label{ocrf:fig:LSAnomaly_roc_pr}
  \centering
  \includegraphics[trim=175 80 175 123, clip, width=0.85\linewidth]{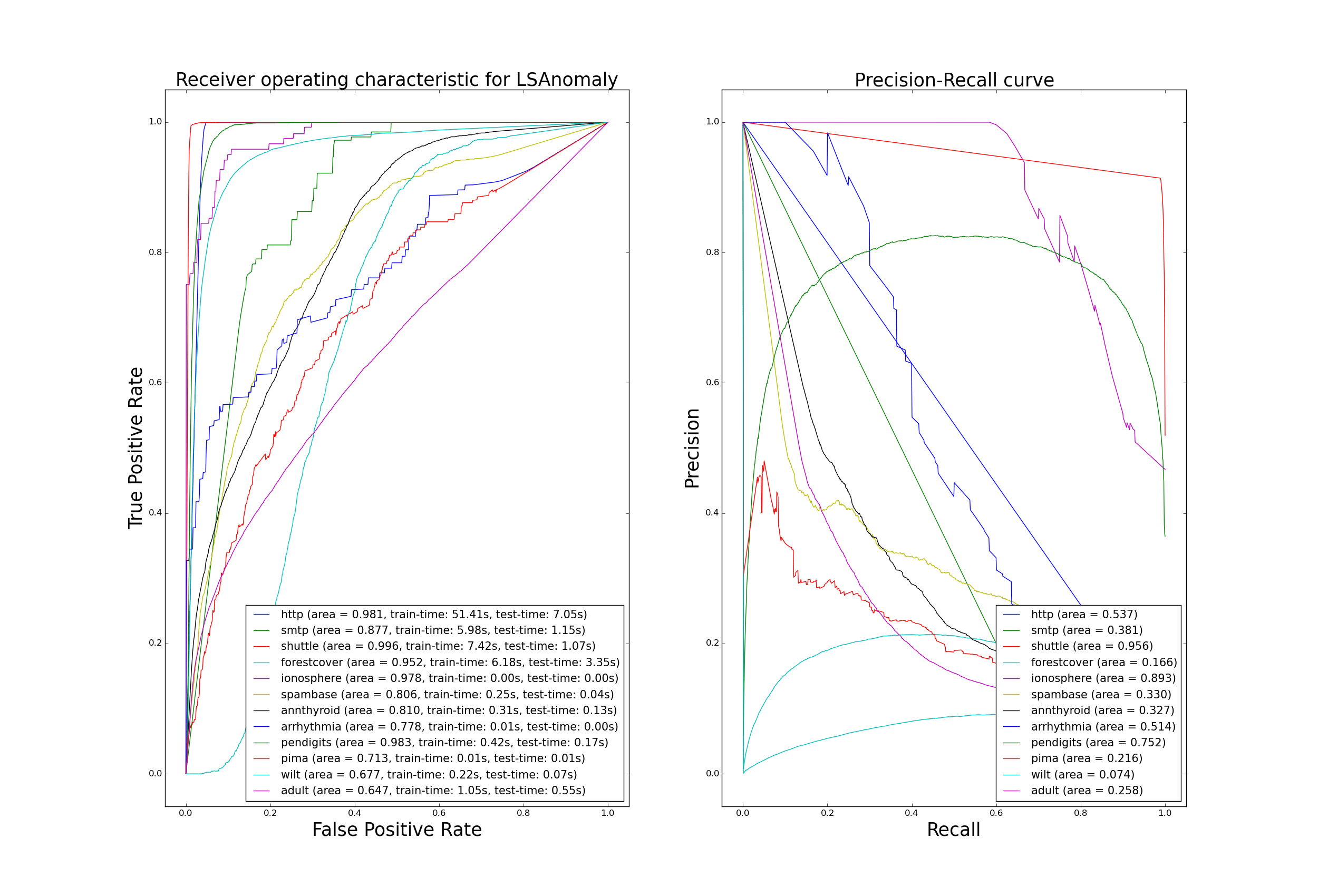}
\end{figure}
\begin{figure}[!ht]
  \caption{ROC and PR curves for LSAD (outlier detection framework)}
  \label{ocrf:fig:LSAnomaly_roc_pr_unsupervised}
  \centering
  \includegraphics[trim=175 80 175 123, clip, width=0.85\linewidth]{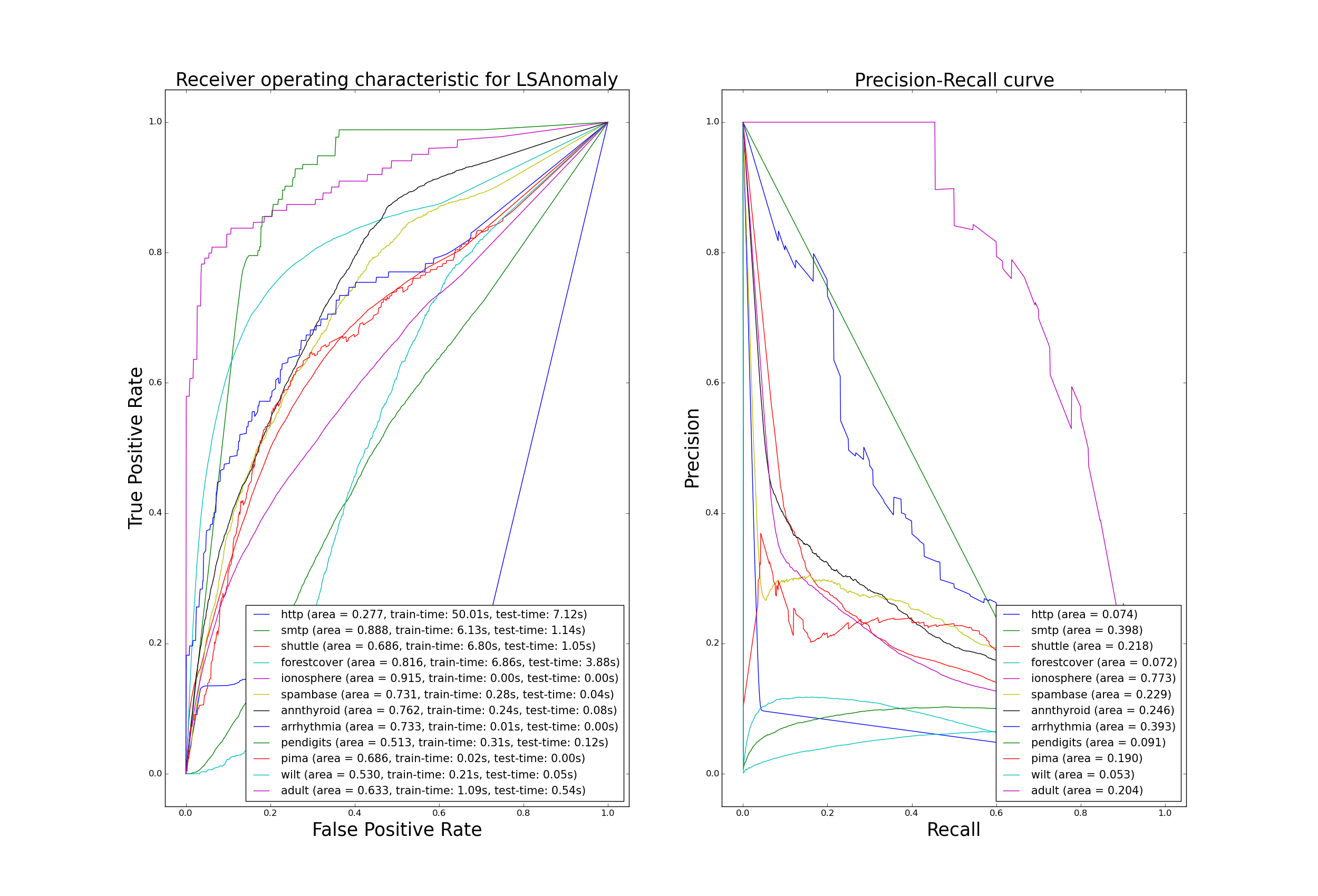}
\end{figure}

\begin{figure}[!ht]
  \caption{ROC and PR curves for RFC (novelty detection framework)}
  \label{ocrf:fig:rf_roc_pr}
  \centering
  \includegraphics[trim=175 80 175 123, clip, width=0.85\linewidth]{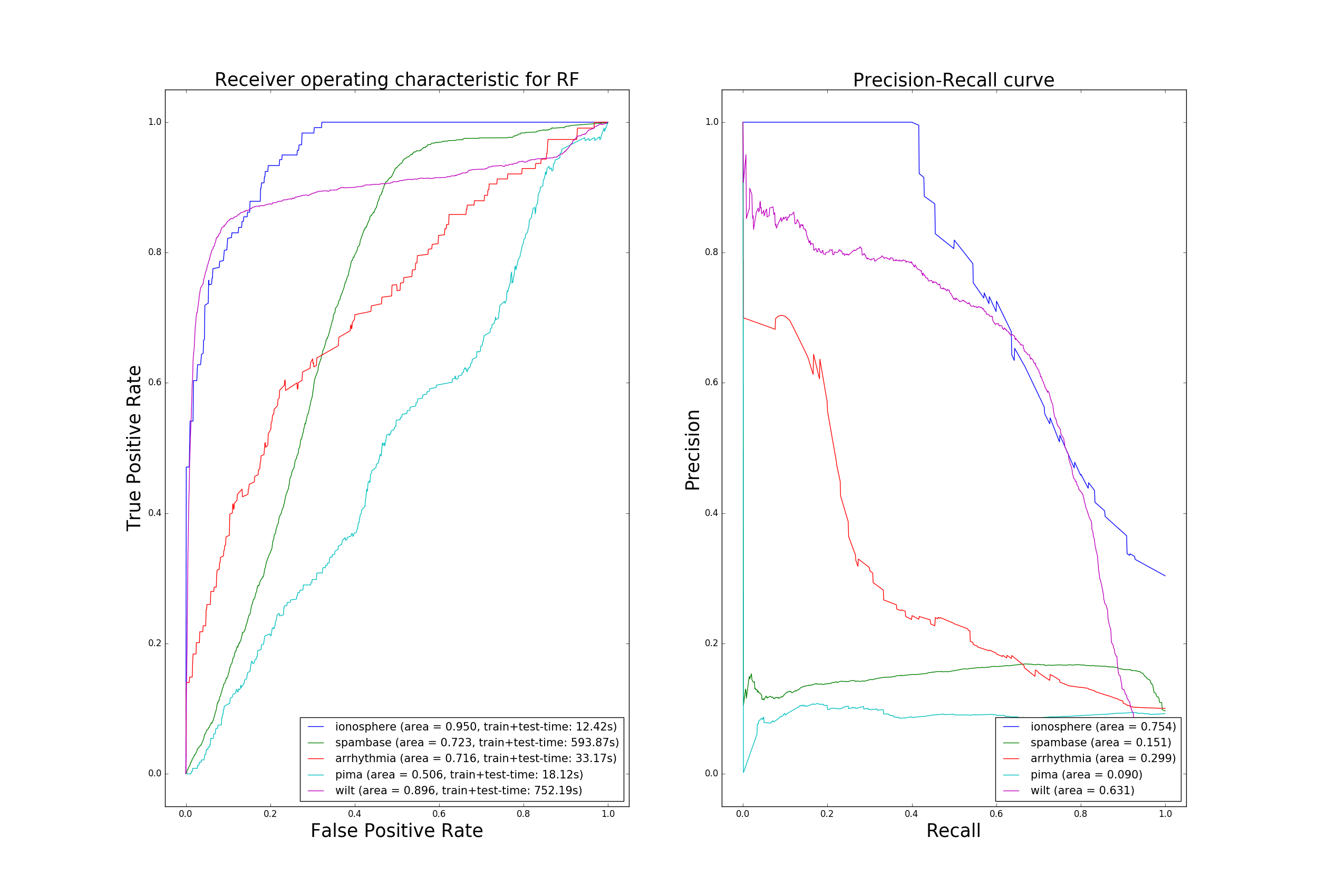}
\end{figure}
\begin{figure}[!ht]
  \caption{ROC and PR curves for RFC (outlier detection framework)}
  \label{ocrf:fig:rf_roc_pr_unsupervised}
  \centering
  \includegraphics[trim=175 80 175 123, clip, width=0.85\linewidth]{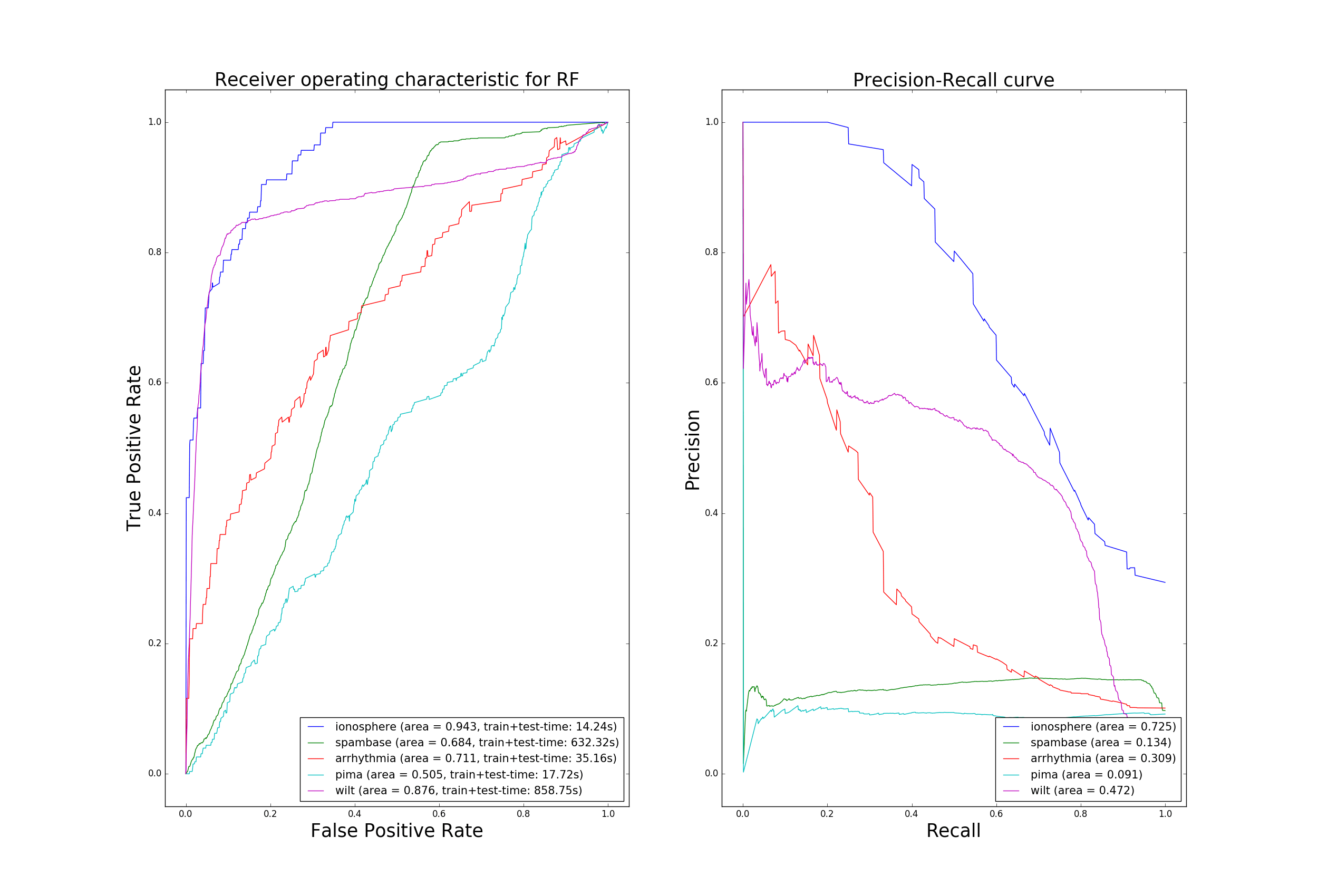}
\end{figure}

\end{document}